\pdftrailerid{}
\PassOptionsToPackage{cmyk}{xcolor}
\documentclass[letterpaper]{article}

\usepackage[preprint]{aaai2027}
\nocopyright

\usepackage{amsmath}
\usepackage{amssymb}
\usepackage{booktabs}
\usepackage{array}
\usepackage{multirow}
\usepackage{longtable}
\usepackage{xurl}
\usepackage{graphicx}
\usepackage{capt-of}
\usepackage{algorithm}
\usepackage{algpseudocode}
\usepackage{afterpage}
\usepackage{natbib}

\makeatletter
\newcommand{\algorithmcaptionfootnotesize}{%
  \def\@fs@capt##1##2{\footnotesize{\@fs@cfont ##1} ##2\par}}
\makeatother

\usepackage{xcolor}
\usepackage[most]{tcolorbox}

\makeatletter
\gdef\pgf@sys@pgf@resource@list@colorspaces{/pgfpcmyk [/Pattern /DeviceCMYK]}
\makeatother

\newtcolorbox{findingbox}{
  enhanced,
  colback=gray!6,
  colframe=black,
  boxrule=0pt,
  borderline west={2pt}{0pt}{black},
  sharp corners,
  left=6pt,
  right=6pt,
  top=5pt,
  bottom=5pt,
  before skip=7pt,
  after skip=7pt
}
\begin{document}
\title{When Teacher Guidance Misleads: Reward-Aligned On-Policy Distillation}
\author{
Siyuan Gan\textsuperscript{\rm 1},
Yuhan Li\textsuperscript{\rm 1,2},
Xiran Wang\textsuperscript{\rm 1,2},
Linjian Meng\corresponding\textsuperscript{\rm 2},
Boyan Wang\corresponding\textsuperscript{\rm 1},\\
Zhen Zhao\textsuperscript{\rm 2},
Jing Huo\textsuperscript{\rm 1},
Yang Gao\textsuperscript{\rm 1}
}
\affiliations{
\textsuperscript{\rm 1}State Key Laboratory of Novel Software Technology, Nanjing University, Nanjing, China\\
\textsuperscript{\rm 2}Shanghai Artificial Intelligence Laboratory, Shanghai, China\\
gansiyuan@smail.nju.edu.cn, menglinjian@pjlab.org.cn, boyanwang@nju.edu.cn
}
\maketitle
\begin{abstract}
  On-policy distillation (OPD) has recently emerged as a popular post-training paradigm for large language models (LLMs), providing an efficient way to transfer the knowledge and capabilities of teacher models into student models. However, teacher guidance on student-generated prefixes is not always reliable. Training should optimize the model to generate responses that are more likely to be correct, or equivalently, to get higher outcome rewards. But during OPD, the teacher model may provide guidance that discourages the student from moving toward correct trajectories or moves the student toward incorrect ones, which is misaligned with outcome reward. Such misaligned guidance is unreliable, as it would mislead the optimization process and ultimately degrade model performance. To mitigate misaligned teacher guidance, we propose \textbf{Reward-Aligned On-Policy Distillation (RA-OPD)}. The key insight is to keep only trajectories whose induced updates move the student toward correct trajectories or discourage the student from moving toward incorrect ones. Specifically, for each sampled trajectory, RA-OPD checks whether its trajectory-level distillation return is consistent with its outcome reward and then filters out the misaligned trajectories. RA-OPD selects more reliable trajectories to improve student model performance without requiring additional computational cost. We evaluate RA-OPD on math and code benchmarks using models from the Qwen3 family and the DeepSeek-R1 family. Across seven math benchmarks and three code benchmarks, RA-OPD significantly outperforms standard OPD and other tested OPD variants.
\end{abstract}

\begin{figure}[t]
\centering
\includegraphics[width=\columnwidth]{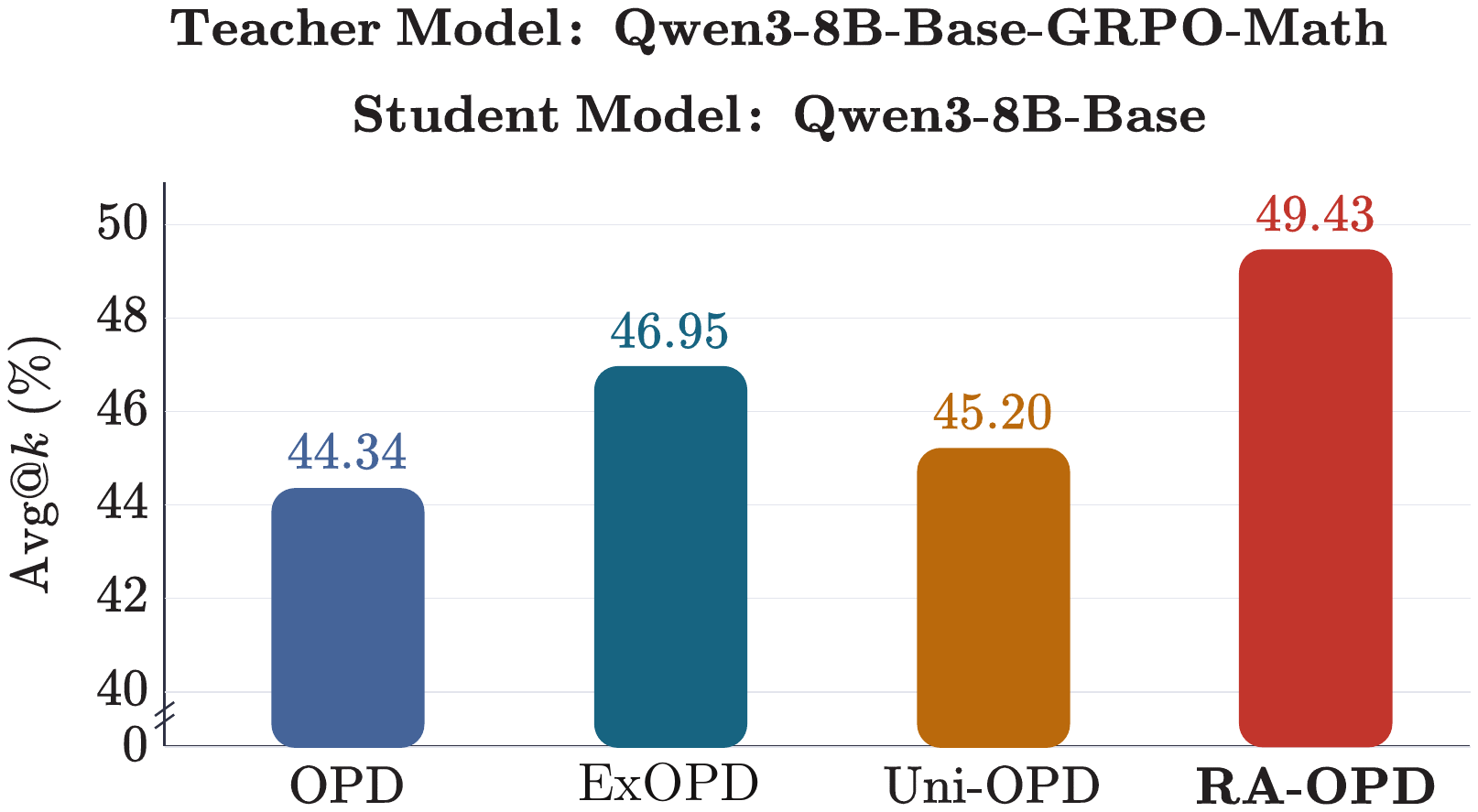}
\vspace{-0.4cm}
\caption{Mean avg@$k$ across seven math reasoning benchmarks for Qwen3-8B-Base distilled from Qwen3-8B-Base-GRPO-Math. RA-OPD achieves the strongest average performance among the tested OPD methods.}
\label{fig:outcome-aligned-performance}
\end{figure}

\section{Introduction}

\begin{figure*}[t]
\centering
\includegraphics[width=\textwidth]{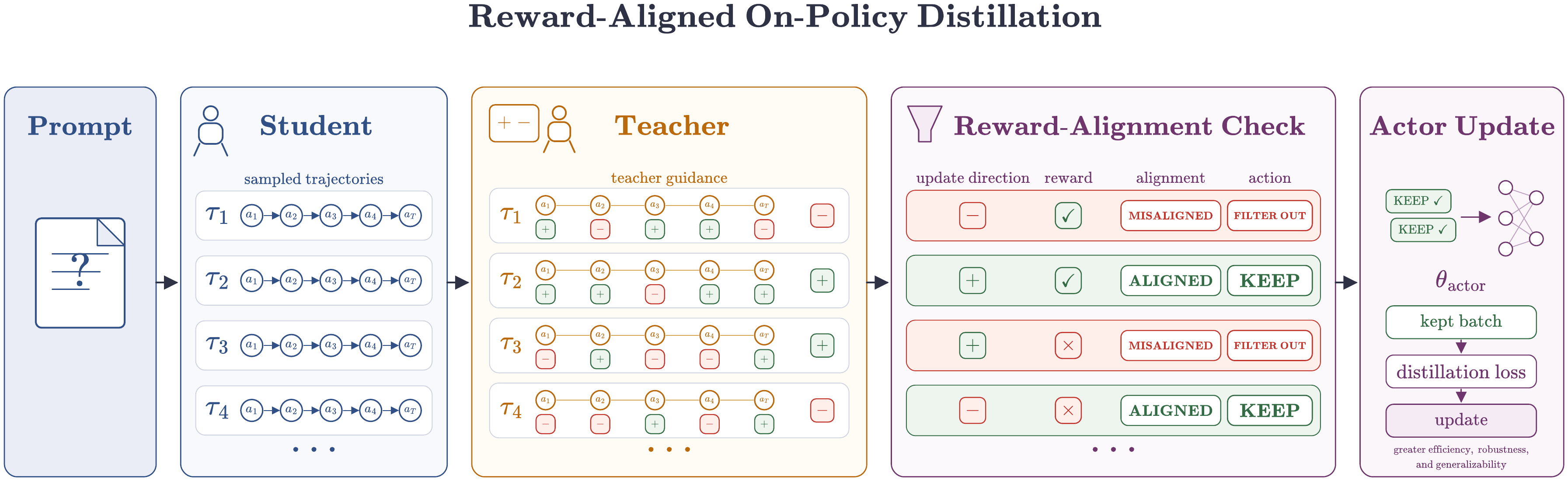}
\caption{RA-OPD detects unreliable teacher guidance by comparing each trajectory's trajectory-level distillation return with its outcome reward and filters out reward-misaligned trajectories before policy optimization.}
\label{fig:ra-opd-overview}
\end{figure*}

On-policy distillation (OPD) has recently emerged as a popular post-training paradigm for large language models (LLMs), providing an efficient way to transfer the knowledge and capabilities of teacher models into student models. Recent frontier models such as Qwen3, MiMo, and GLM-5 have adopted OPD in their post-training pipelines~\cite{yang2025qwen3,xiao2026mimo,zeng2026glm5}. Unlike off-policy distillation, which trains the student on fixed teacher-generated prefixes, OPD trains the student on student-generated prefixes with teacher guidance under a reverse KL objective. Consequently, OPD alleviates the training-inference prefix mismatch, thereby reducing exposure bias and compounding errors under student-generated prefixes, which degrades model performance~\cite{agarwal2024onpolicy}.

Unfortunately, teacher guidance on student-generated prefixes is not always reliable during OPD, constituting a key factor behind performance degradation and even training failure~\cite{li2026rethinking,song2026survey}. Specifically, for verifiable tasks such as math and code, training should optimize the model to generate trajectories that are more likely to be correct, or equivalently, to get higher outcome rewards. But the teacher model may provide guidance that discourages the student from moving toward correct trajectories or moves the student toward incorrect ones, which is misaligned with outcome reward. Such misaligned guidance is unreliable, as it would mislead the optimization process and ultimately degrade model performance~\cite{li2026rethinking,song2026survey,hou2026uniopd}. 

To get more reliable teacher guidance, several works improve the reliability of teacher guidanceat the token-level by restricting, reweighting, or truncating teacher guidance~\cite{zhou2026less,liu2026teacher,yang2026pruneopd}. However, these methods assess teacher guidance primarily using local, token-level guidance, so the teacher model may also provide misaligned trajectory-level teacher guidance~\cite{zhou2026less,liu2026teacher,yang2026pruneopd,jiang2026trajectory}. Furthermore, Uni-OPD aggregates token-level teacher guidance into a trajectory-level distillation return and then ensures relative consistency between the distillation returns and the outcome rewards~\cite{hou2026uniopd}. However, its reliability criterion has two limitations. First, its relative consistency requires multiple rollouts for each prompt, increasing computational cost. Second, relative consistency cannot be established for prompts whose sampled trajectories share the same reward, so Uni-OPD falls back to the original OPD update for those prompts.

To address these limitations in prior work, we propose \textbf{Reward-Aligned On-Policy Distillation (RA-OPD)}. The key insight of RA-OPD is to keep only trajectories whose induced updates move the student toward correct trajectories or discourage the student from moving toward incorrect ones. Specifically, for each sampled trajectory, RA-OPD checks whether its trajectory-level distillation return is consistent with its outcome reward and then filters out the misaligned trajectories. First, RA-OPD operates on the student trajectories already generated for OPD and reuses the corresponding teacher probabilities, thereby requiring no additional student rollouts or teacher evaluations. Moreover, RA-OPD makes an absolute reliability decision for each trajectory independently. The criterion remains well defined with a single trajectory per prompt and does not require the reward diversity of the current student policy.

We evaluate RA-OPD on math and code benchmarks using models from the Qwen3 family~\cite{yang2025qwen3} and the DeepSeek-R1 family~\cite{guo2025deepseekr1}. Across seven math benchmarks and three code benchmarks, RA-OPD significantly outperforms other tested OPD methods.

\section{Preliminaries}

\subsection{On-Policy Distillation}

We first introduce standard OPD and the teacher guidance used in the subsequent trajectory-level formulation.

First, let $q\sim\mathcal{D}$ denote an input question, and let $\pi_\theta$ and $\pi_T$ denote the student and teacher policies, respectively.  The student samples a trajectory $\tau=(o_1,\ldots,o_{|\tau|})\sim\pi_\theta(\cdot\mid q)$, where $o_{<t}=(o_1,\ldots,o_{t-1})$ is the response prefix before position $t$. Let $d_{\mathrm{roll}}$ denote the distribution over response prefixes $s_t=(q,o_{<t})$ visited by the current student rollouts.  Under the reverse KL objective considered in this work, OPD minimizes
\begin{equation*}
J_{\mathrm{OPD}}(\theta)
=
\mathbb{E}_{s_t\sim d_{\mathrm{roll}}}
\left[
D_{\mathrm{KL}}
\bigl(
\pi_\theta(\cdot\mid s_t)
\,\|\,
\pi_T(\cdot\mid s_t)
\bigr)
\right].
\end{equation*}
During each policy update, the sampled prefixes in the rollout batch are treated as fixed.  Accordingly, for a student-generated token $o_t$, the corresponding teacher guidance is~\cite{schulman2017ppo,oh2026vopd}
\begin{equation*}
r_t^{\mathrm{OPD}}(q,\tau)
=
\log
\frac{
\pi_T(o_t\mid q,o_{<t})
}{
\pi_\theta(o_t\mid q,o_{<t})
}.
\end{equation*}
Next, for a rollout batch $\mathcal{B}$ with $N=\sum_{(q,\tau)\in\mathcal{B}}|\tau|$ valid response tokens, the OPD policy-gradient surrogate is
\begin{equation*}
\widehat{\mathcal{L}}_{\mathrm{OPD}}(\theta)
=
-\frac{1}{N}
\sum_{(q,\tau)\in\mathcal{B}}
\sum_{t=1}^{|\tau|}
\operatorname{sg}\!\left[r_t^{\mathrm{OPD}}\right]
\log\pi_\theta(o_t\mid s_t).
\end{equation*}
Here, $\operatorname{sg}[\cdot]$ denotes stop-gradient.  Thus, the guidance is computed from the sampled trajectory and treated as a fixed training signal. Consequently, a positive value would increase the probability of the sampled token, whereas a negative value would decrease it.

\subsection{Uni-OPD}

Building on standard OPD, Uni-OPD improves OPD from both the student and teacher perspectives~\cite{hou2026uniopd}.  On the student side, it uses offline and online data balancing to promote exploration.  On the teacher side, it introduces outcome-guided margin calibration based on the guidance aggregated over a complete trajectory.  Since our method also builds on this characterization of teacher guidance, we focus on the teacher-side formulation below.

Specifically, for each trajectory, Uni-OPD defines the trajectory-level distillation return
\begin{equation}
G_{\mathrm{OPD}}\!(q,\tau)
\!=\!
\frac{1}{|\tau|}\!
\sum_{\!t=1}^{|\tau|}
r_t^{\mathrm{OPD}}\!(q,\tau)
\!=\!
\frac{1}{|\tau|}\!
\log\!
\frac{\pi_T(\tau\mid q)}
{\pi_\theta(\tau\mid q)}.
\label{eq:trajectory-guidance}
\end{equation}
Here, the length normalization makes the returns comparable across trajectories. Under autoregressive factorization, this return is the log-ratio between the probabilities assigned to the complete trajectory by the teacher and student. Its sign indicates whether the teacher assigns a higher or lower probability to the trajectory than the student.

Next, given the binary outcome reward $R(q,\tau)\in\{0,1\}$, Uni-OPD partitions the trajectories sampled for the same question into the following outcome-specific sets
\begin{gather*}
S_{+}(q)=\{\tau\mid R(q,\tau)=1\},\\
S_{-}(q)=\{\tau\mid R(q,\tau)=0\}.
\end{gather*}
Finally, for a rollout group containing both outcomes, Uni-OPD requires
\begin{equation*}
\min_{\tau\in S_{+}(q)}
G_{\mathrm{OPD}}(q,\tau)
\geq
\max_{\tau\in S_{-}(q)}
G_{\mathrm{OPD}}(q,\tau)
+
\delta,
\end{equation*}
where $\delta>0$ is a safety margin.  Uni-OPD enforces this condition through either margin masking, which discards groups that violate it, or margin shifting, which adjusts their returns to restore the required separation.

Nevertheless, the relative consistency criterion depends on multiple rollouts for each prompt and therefore increases computational cost.  Moreover, when the sampled trajectories share the same reward, relative consistency cannot be established and Uni-OPD falls back to the original OPD.  Overall, Uni-OPD introduces additional rollout overhead and cannot reliably assess teacher guidance for single-rollout or reward-homogeneous prompt groups.

\section{Method}

To obtain more reliable teacher guidance and address the limitations of prior work, we propose \textbf{Reward-Aligned On-Policy Distillation (RA-OPD)}. RA-OPD aggregates token-level teacher guidance into a trajectory-level distillation return, compares this return with the verified outcome reward, and filters out reward-misaligned trajectories. The complete implementation is provided in Algorithm~\ref{alg:raopd}.

\subsection{Motivation}

To summarize teacher guidance at trajectory-level, we aggregate the OPD guidance into the trajectory-level distillation return $G_{\mathrm{OPD}}$, as defined in Eq.~\eqref{eq:trajectory-guidance}. Reliable teacher guidance in OPD should be consistent with final outcome reward. It should move the student toward correct trajectories and discourage the student from moving toward incorrect ones.

However, this alignment is not guaranteed in practice, as demonstrated by the substantial fraction of reward-misaligned trajectories observed throughout training in Figure~\ref{fig:drop-fractions}. Specifically, local patterns in an incorrect trajectory may fall into the teacher's high-confidence regions and produce a positive return, whereas a correct trajectory that deviates from the teacher's dominant patterns may produce a negative return~\cite{armandpour2026unmasking,li2026rethinking,song2026survey}. In the former case, OPD moves the student toward an incorrect trajectory; in the latter, it discourages the student from moving toward a correct one. Both cases constitute misaligned teacher guidance.

These cases expose a limitation of the relative consistency criterion used in prior work. Comparing trajectories within a prompt does not determine whether the distillation return for each trajectory is consistent with its own outcome reward. The key insight of RA-OPD is to use the verified outcome reward as a trajectory-level reliability criterion and filter out trajectories whose distillation returns are inconsistent with their outcome rewards before policy optimization.

\subsection{Details of RA-OPD}

\begin{algorithm}[t]
\algorithmcaptionfootnotesize
\footnotesize
\caption{Reward-Aligned On-Policy Distillation}
\label{alg:raopd}
\begin{algorithmic}[1]
\Require Student $\pi_\theta$, teacher $\pi_T$, verifier $\operatorname{Verify}$, dataset $\mathcal{D}$
\For{each training iteration}
    \State Sample $\mathcal{B}=\{(q_i,\tau_i)\}_{i=1}^{B}$
    \Statex \hspace{\algorithmicindent} where $q_i\sim\mathcal{D}$ and $\tau_i\sim\pi_\theta(\cdot\mid q_i)$
    \For{each $(q_i,\tau_i)\in\mathcal{B}$}
        \For{$t=1,\ldots,|\tau_i|$}
            \State Set $s_{i,t}=(q_i,o_{i,<t})$
            \State $r_{i,t}^{\mathrm{OPD}}=\log\pi_T(o_{i,t}\mid s_{i,t})$
            \Statex \hspace{\algorithmicindent}\hspace{\algorithmicindent}
            $\displaystyle\phantom{r_{i,t}^{\mathrm{OPD}}={}}-\log\pi_\theta(o_{i,t}\mid s_{i,t})$
        \EndFor
        \State Compute $G_i=|\tau_i|^{-1}\sum_{t=1}^{|\tau_i|}r_{i,t}^{\mathrm{OPD}}$
        \State Obtain $R_i=\operatorname{Verify}(q_i,\tau_i)\in\{0,1\}$
        \State Compute $m_i=\mathbf{1}[(2R_i-1)G_i\geq 0]$
    \EndFor
    \State Compute $Z=\sum_{i=1}^{B}m_i|\tau_i|$
    \If{$Z>0$}
        \State Minimize $\mathcal{L}_{\mathrm{RA\text{-}OPD}}$ in Eq.~\eqref{eq:raopd-loss}
    \Else
        \State Set the distillation gradient to zero
    \EndIf
\EndFor
\end{algorithmic}
\end{algorithm}

RA-OPD applies a reward-alignment check to every student-generated trajectory before policy optimization. For each trajectory $(q_i,\tau_i)$, let $G_i=G_{\mathrm{OPD}}(q_i,\tau_i)$ denote the trajectory-level distillation return defined in Eq.~\eqref{eq:trajectory-guidance}. We obtain its binary outcome reward from an outcome verifier:
\begin{equation*}
R_i
=
\operatorname{Verify}(q_i,\tau_i)
\in\{0,1\},
\end{equation*}
where $R_i=1$ indicates that the final answer is correct and $R_i=0$ otherwise. We then define the binary reward-alignment mask as
\begin{equation}
m_i
=
\mathbf{1}
\left[
(2R_i-1)G_i\geq 0
\right].
\label{eq:ra-mask}
\end{equation}
A trajectory is kept when $m_i=1$ and filtered out when $m_i=0$. Equivalently, RA-OPD keeps a correct trajectory when its distillation return is non-negative and keeps an incorrect trajectory when its distillation return is non-positive. We regard $G_i=0$ as a neutral boundary and keep the trajectory because it does not indicate a conflicting teacher preference. Since the positive length normalization in $G_i$ does not change its sign, the filtering decision is also unchanged if the unnormalized sum of token-level OPD rewards is used.

Consider a standard OPD rollout batch $\mathcal{B}=\{(q_i,\tau_i)\}_{i=1}^{B}$. RA-OPD applies $m_i$ to every valid response token in trajectory $\tau_i$. Let
\begin{equation*}
Z
=
\sum_{i=1}^{B}m_i|\tau_i|
\end{equation*}
denote the number of valid response tokens in the kept trajectories. When $Z>0$, the filtered OPD policy-gradient surrogate over the kept trajectories is
\begin{equation}
\scalebox{0.90}[1]{\ensuremath{\displaystyle
\mathcal{L}_{\mathrm{RA\!-\!OPD}}\!(\theta)
\!=\!
-\frac{1}{Z}\!
\sum_{\!i=1}^{B}m_i\!
\sum_{\!t=1}^{|\tau_i|}
\operatorname{sg}\!\left[r_{i,t}^{\mathrm{OPD}}\right]\!
\log p_{\theta,i,t}\!(o_{i,t}),
}}
\label{eq:raopd-loss}
\end{equation}
where $p_{\theta,i,t}$ is the student next-token distribution on prefix $(q_i,o_{i,<t})$, and $\operatorname{sg}[\cdot]$ denotes stop-gradient. The OPD rewards and the reward-alignment mask are computed once from the rollout batch and treated as fixed training signals; gradients are not propagated through them. A positive $r_{i,t}^{\mathrm{OPD}}$ therefore increases the log-probability of the sampled token under gradient descent, whereas a negative value decreases it. The mask only determines whether the complete trajectory contributes to optimization; it does not modify the token-level OPD rewards within a kept trajectory. Thus, RA-OPD preserves the standard OPD objective on the kept samples. When $Z=0$, the RA-OPD distillation gradient is zero.

\subsection{Discussion}

Compared with standard OPD and existing methods for improving its reliability, RA-OPD offers several practical advantages. First, it operates on the student trajectories already generated for OPD and reuses the corresponding teacher probabilities, thereby requiring no additional student rollouts or teacher evaluations. Second, RA-OPD makes an absolute reliability decision for each trajectory independently. The criterion remains well defined with a single trajectory per prompt and does not require the reward diversity of the current student policy. Third, RA-OPD is simple, requiring only the binary consistency mask in Eq.~\eqref{eq:ra-mask} to determine whether a trajectory contributes to standard OPD optimization. Furthermore, by the outcome verifier, RA-OPD is interpretable because every filtered trajectory can be attributed to one of two explicit conflicts: either discouraging the student from moving toward a correct response or moving the student toward an incorrect response.

RA-OPD is also modular and can be combined with other methods for improving OPD reliability. Methods such as Early Stopping Rollout, Lookahead Group Reward, and Prune-OPD restrict, reweight, or truncate local teacher guidance~\cite{zhou2026less,liu2026teacher,yang2026pruneopd}. For these methods, the effective token-level signals produced by the base method can first be aggregated into a trajectory-level distillation return, after which the RA-OPD mask can be applied. This modularity allows reward-aligned filtering to strengthen an existing OPD pipeline while preserving the base method's underlying optimization objective.

\section{Experiments}
\label{sec:experiments}

\subsection{Experimental Setup}
\label{subsec:experimental-setup}

\begin{table*}[t]
\centering
\begin{tabular*}{\textwidth}{@{\extracolsep{\fill}}cl*{8}{c}@{}}
\toprule
\multicolumn{10}{c}{\textbf{avg@$k$}} \\
\midrule
\textbf{Student} & \multicolumn{1}{c}{\textbf{Method}}
& \textbf{AIME24} & \textbf{AIME25} & \textbf{AIME26}
& \textbf{AMC} & \textbf{MATH} & \textbf{Minerva}
& \textbf{Olympiad} & \textbf{Avg.} \\
\midrule
\multirow{5}{*}{Qwen3-4B-Base}
& Base Model & 8.02 & 5.31 & 6.15 & 30.80 & 54.07 & 22.79 & 25.22 & 21.77 \\
& OPD       & 21.04 & 19.17 & 14.27 & 57.83 & \underline{84.67} & 34.33 & 53.44 & 40.68 \\
& ExOPD     & \underline{25.73} & \underline{20.42} & 18.65 & \underline{58.81} & 83.65 & 36.63 & 52.63 & \underline{42.36} \\
& Uni-OPD   & 21.67 & 18.12 & \underline{18.85} & \underline{58.81} & 83.88 & \underline{37.04} & \underline{53.52} & 41.70 \\
& \textbf{RA-OPD}
& \textbf{27.08} & \textbf{23.75} & \textbf{21.77}
& \textbf{65.36} & \textbf{87.98} & \textbf{39.71}
& \textbf{55.52} & \textbf{45.88} \\
\midrule
\multirow{5}{*}{Qwen3-8B-Base}
& Base Model & 7.19 & 12.08 & 7.71 & 37.24 & 62.98 & 25.74 & 31.19 & 26.30 \\
& OPD       & 25.00 & 21.88 & 18.85 & 65.55 & 88.18 & 33.87 & 57.04 & 44.34 \\
& ExOPD     & \underline{27.50} & \underline{24.79} & \textbf{28.12} & \underline{66.15} & \underline{88.90} & 35.89 & \underline{57.28} & \underline{46.95} \\
& Uni-OPD   & 27.29 & 21.88 & 22.29 & 63.25 & 88.12 & \underline{36.76} & 56.78 & 45.20 \\
& \textbf{RA-OPD}
& \textbf{33.75} & \textbf{25.83} & \underline{27.60}
& \textbf{71.50} & \textbf{89.35} & \textbf{40.49}
& \textbf{57.50} & \textbf{49.43} \\
\midrule
\multicolumn{10}{c}{\textbf{pass@$k$}} \\
\midrule
\textbf{Student} & \multicolumn{1}{c}{\textbf{Method}}
& \textbf{AIME24} & \textbf{AIME25} & \textbf{AIME26}
& \textbf{AMC} & \textbf{MATH} & \textbf{Minerva}
& \textbf{Olympiad} & \textbf{Avg.} \\
\midrule
\multirow{5}{*}{Qwen3-4B-Base}
& Base Model & 26.67 & 26.67 & 20.00 & 63.86 & 76.20 & 36.03 & 42.37 & 41.69 \\
& OPD       & 40.00 & \underline{40.00} & 33.33 & \textbf{79.52} & \textbf{90.40} & 42.65 & \underline{63.70} & 55.66 \\
& ExOPD     & 46.67 & \textbf{46.67} & \textbf{46.67} & \underline{77.11} & 84.40 & 40.07 & 54.22 & 56.54 \\
& Uni-OPD   & \underline{50.00} & 36.67 & \underline{43.33} & \textbf{79.52} & 85.00 & \underline{45.22} & 57.78 & \underline{56.79} \\
& \textbf{RA-OPD}
& \textbf{53.33} & \textbf{46.67} & \textbf{46.67}
& \textbf{79.52} & \underline{89.20} & \textbf{48.53}
& \textbf{63.85} & \textbf{61.11} \\
\midrule
\multirow{5}{*}{Qwen3-8B-Base}
& Base Model & 33.33 & 33.33 & 23.33 & 75.90 & 82.40 & 37.50 & 47.70 & 47.64 \\
& OPD       & \underline{46.67} & \underline{40.00} & 30.00 & \underline{86.75} & \textbf{93.80} & 39.71 & 66.67 & 57.66 \\
& ExOPD     & 40.00 & 36.67 & \textbf{46.67} & 85.54 & 92.20 & 39.71 & \textbf{68.74} & \underline{58.50} \\
& Uni-OPD   & \textbf{50.00} & 36.67 & 40.00 & 83.13 & 91.60 & \underline{41.54} & 66.37 & 58.47 \\
& \textbf{RA-OPD}
& \underline{46.67} & \textbf{46.67} & \underline{43.33}
& \textbf{90.36} & \underline{93.00} & \textbf{43.75}
& \underline{67.11} & \textbf{61.56} \\
\bottomrule
\end{tabular*}
\caption{Math reasoning results (\%) for the Qwen3 family.  We report avg@$k$ and pass@$k$, with $k=32$ for AIME and AMC and $k=8$ for all other benchmarks.  Within each student block, the best and second-best results are shown in bold and underlined, respectively.}
\label{tab:main-math}
\end{table*}

\begin{table*}[t]
\centering
\begin{tabular*}{\textwidth}{@{\extracolsep{\fill}}cl*{8}{c}@{}}
\toprule
\multicolumn{10}{c}{\textbf{avg@$k$}} \\
\midrule
\textbf{Student} & \multicolumn{1}{c}{\textbf{Method}}
& \textbf{AIME24} & \textbf{AIME25} & \textbf{AIME26}
& \textbf{AMC} & \textbf{MATH} & \textbf{Minerva}
& \textbf{Olympiad} & \textbf{Avg.} \\
\midrule
\multirow{5}{*}{\shortstack{DeepSeek-R1-\\Distill-Qwen-7B}}
& Base Model & 54.58 & 40.00 & 47.50 & 82.83 & 94.27 & 37.50 & 67.15 & 60.55 \\
& OPD        & 57.19 & 46.15 & 55.31 & 86.60 & 95.15 & 38.51 & 72.09 & 64.43 \\
& ExOPD      & 59.79 & 48.75 & 57.29 & \underline{86.94} & 95.43 & \underline{38.79} & 71.85 & 65.55 \\
& Uni-OPD    & \underline{60.83} & \underline{51.77} & \underline{60.21} & 86.07 & \underline{95.55} & \underline{38.79} & \underline{72.65} & \underline{66.55} \\
& \textbf{RA-OPD}
& \textbf{65.62} & \textbf{55.31} & \textbf{64.17}
& \textbf{88.70} & \textbf{96.53} & \textbf{41.13}
& \textbf{73.93} & \textbf{69.34} \\
\midrule
\multicolumn{10}{c}{\textbf{pass@$k$}} \\
\midrule
\textbf{Student} & \multicolumn{1}{c}{\textbf{Method}}
& \textbf{AIME24} & \textbf{AIME25} & \textbf{AIME26}
& \textbf{AMC} & \textbf{MATH} & \textbf{Minerva}
& \textbf{Olympiad} & \textbf{Avg.} \\
\midrule
\multirow{5}{*}{\shortstack{DeepSeek-R1-\\Distill-Qwen-7B}}
& Base Model & \textbf{86.67} & 73.33 & 76.67 & \textbf{98.80} & \underline{99.00} & \underline{48.16} & 82.52 & 80.74 \\
& OPD        & \underline{83.33} & \underline{76.67} & 80.00 & \underline{97.59} & 98.80 & 47.06 & 82.22 & \underline{80.81} \\
& ExOPD      & \underline{83.33} & 66.67 & 76.67 & \underline{97.59} & \underline{99.00} & 45.59 & \underline{83.85} & 78.96 \\
& Uni-OPD    & 80.00 & 73.33 & \underline{83.33} & \underline{97.59} & \underline{99.00} & 44.49 & 81.93 & 79.95 \\
& \textbf{RA-OPD}
& \textbf{86.67} & \textbf{80.00} & \textbf{86.67}
& \underline{97.59} & \textbf{99.60} & \textbf{49.26}
& \textbf{84.59} & \textbf{83.48} \\
\bottomrule
\end{tabular*}
\caption{Math reasoning results (\%) for the DeepSeek model series, with Skywork-OR1-Math-7B as the teacher and DeepSeek-R1-Distill-Qwen-7B as the student.  Results follow the avg@$k$ and pass@$k$ evaluation protocol of Table~\ref{tab:main-math}; the best and second-best results are shown in bold and underlined, respectively.}
\label{tab:deepseek-results}
\end{table*}

\afterpage{\afterpage{\afterpage{\afterpage{%
\begin{figure}[!t]
\centering
\includegraphics[width=0.96\columnwidth]{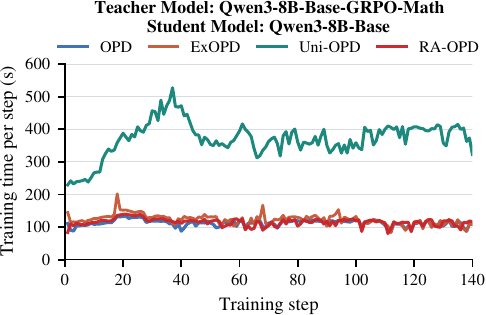}
\caption{Per-step training time for all tested OPD methods.}
\label{fig:training-time}
\end{figure}
}}}}

\afterpage{\afterpage{\afterpage{\afterpage{%
\begin{figure*}[!t]
\centering
\includegraphics[width=0.98\textwidth]{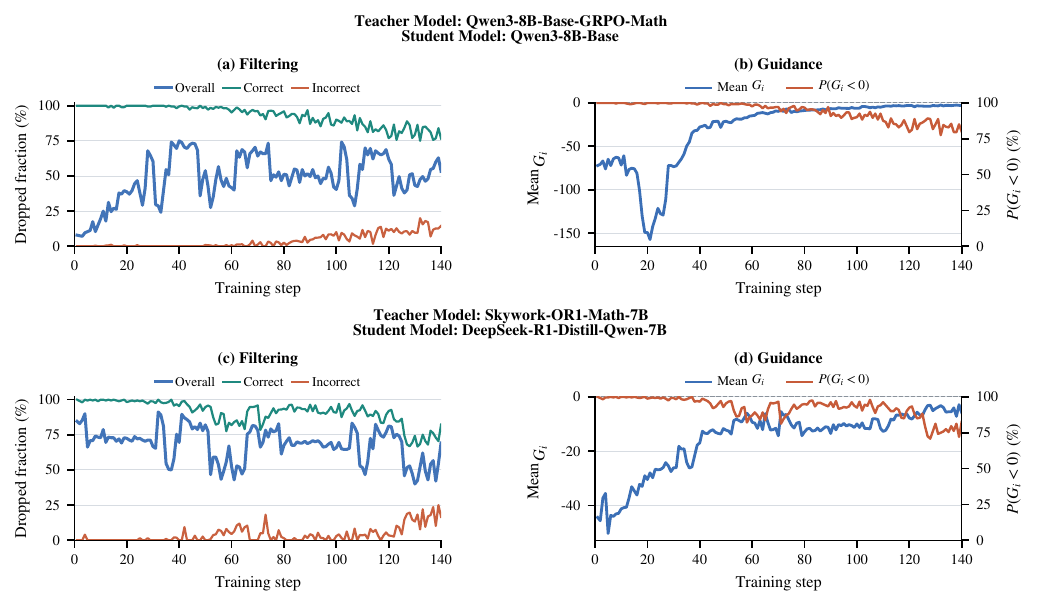}
\caption{Training dynamics of reward-aligned filtering and trajectory-level distillation returns.  Panels (a)--(b) show Qwen3-8B-Base-GRPO-Math distilled into Qwen3-8B-Base, while panels (c)--(d) show Skywork-OR1-Math-7B distilled into DeepSeek-R1-Distill-Qwen-7B.  The filtering panels report the overall, correct, and incorrect dropped fractions, whereas the return panels report mean $G_i$ on the left axis and $P(G_i<0)$ on the right axis.}
\label{fig:drop-fractions}
\end{figure*}
}}}}

\paragraph{Models and training data.} For math, DAPO-Math-17K is used to train the Qwen3-8B-Base-GRPO-Math teacher with GRPO~\cite{shao2024deepseekmath} and distill it into Qwen3-4B/8B-Base students~\cite{yang2025qwen3}. To evaluate RA-OPD beyond the Qwen3 family, we directly use the publicly released Skywork-OR1-Math-7B~\cite{he2025skyworkor1} as the teacher, which is developed by post-training DeepSeek-R1-Distill-Qwen-7B. And we use DeepSeek-R1-Distill-Qwen-7B~\cite{guo2025deepseekr1} as the student, with DAPO-Math-17K for OPD. For code, Eurus-2-RL-Data-Code-25K is used for both GRPO teacher training and subsequent OPD from Qwen3-8B-Base-GRPO-Code to Qwen3-4B-Base (\S\ref{subsec:code-results}). We train the student model for two epochs in all OPD experiments.

\paragraph{Baselines.} We compare RA-OPD with the student base model, standard OPD, ExOPD, the reward-extrapolation variant of Generalized OPD (G-OPD)~\cite{yang2026beyond}, and Uni-OPD~\cite{hou2026uniopd}.  For ExOPD and Uni-OPD, we follow the configurations recommended in their respective original papers; all remaining training and rollout hyperparameters are held fixed across methods.

\paragraph{Benchmarks and metrics.} We report avg@$k$ and pass@$k$ for all evaluations.  For math, we evaluate on AIME24/25/26 and AMC with $k=32$, and on MATH~\cite{hendrycks2021math}, Minerva Math~\cite{lewkowycz2022minerva}, and OlympiadBench~\cite{he2024olympiadbench} with $k=8$.  For code, we evaluate on HumanEval$+$ and MBPP$+$~\cite{chen2021evaluating,austin2021program,liu2023evalplus}, and LiveCodeBench v6~\cite{jain2024livecodebench} with $k=4$.

\paragraph{Experimental settings.} Detailed training and evaluation parameter settings are provided in Appendix~\ref{app:training-parameters}.

\subsection{Main Results}
\label{subsec:main-results}

\begin{findingbox}
\textbf{Finding 1.} RA-OPD demonstrates the strongest average math reasoning performance for both Qwen3-4B-Base and Qwen3-8B-Base students.
\end{findingbox}

Table~\ref{tab:main-math} reports the main results. RA-OPD achieves the best average avg@$k$ for both students. On Qwen3-4B-Base, RA-OPD reaches an average avg@$k$ of 45.88, improving standard OPD by 5.20 points and the strongest competing baseline, ExOPD, by 3.52 points.  More specifically, RA-OPD scores 27.08, 23.75, and 21.77 on AIME24, AIME25, and AIME26, improving over OPD by 6.04, 4.58, and 7.50 points, respectively. It also reaches 65.36 on AMC, 87.98 on MATH, 39.71 on Minerva Math, and 55.52 on OlympiadBench (+2.08). The same pattern holds when the student is scaled to Qwen3-8B-Base. RA-OPD leads all competing methods on six of the seven benchmarks. RA-OPD achieves 49.43 average accuracy, 5.09 points above OPD and 2.48 points above ExOPD. Relative to OPD, the scores of 33.75, 25.83, and 27.60 on AIME24, AIME25, and AIME26 correspond to gains of 8.75, 3.95, and 8.75 points, respectively.  The remaining benchmarks exhibit the same improvement, with scores of 71.50 on AMC, 89.35 on MATH, 40.49 on Minerva Math, and 57.50 on OlympiadBench. Only on AIME26, it remains 0.52 points below ExOPD despite its large improvement over standard OPD. Moreover, RA-OPD reaches average pass@$k$ scores of 61.11 and 61.56 for two students, respectively. These results exceed the strongest competing baselines by 4.32 and 3.06 points and confirm that RA-OPD improves model performance coverage across both student model sizes.

\afterpage{\afterpage{\afterpage{%
\begin{table}[!b]
\centering
\begin{tabular*}{\columnwidth}{@{\extracolsep{\fill}}l*{4}{c}@{}}
\toprule
\multicolumn{5}{c}{\textbf{avg@4}} \\
\midrule
\multicolumn{1}{c}{\textbf{Method}}
& \textbf{HE$+$} & \textbf{MBPP$+$}
& \textbf{LCB v6} & \textbf{Avg.} \\
\midrule
Base Model & 18.14 & 13.16 & 16.14 & 15.81 \\
OPD        & 72.10 & 68.72 & 24.29 & 55.04 \\
Uni-OPD    & \underline{75.76} & 66.07 & 23.86 & 55.23 \\
ExOPD      & 74.24 & \underline{68.78} & \underline{25.29} & \underline{56.10} \\
\textbf{RA-OPD}
& \textbf{78.81} & \textbf{70.97} & \textbf{27.14}
& \textbf{58.97} \\
\midrule
\multicolumn{5}{c}{\textbf{pass@4}} \\
\midrule
\multicolumn{1}{c}{\textbf{Method}}
& \textbf{HE$+$} & \textbf{MBPP$+$}
& \textbf{LCB v6} & \textbf{Avg.} \\
\midrule
Base Model & 48.78 & 41.27 & \underline{28.00} & 39.35 \\
OPD        & 84.15 & \textbf{78.84} & \underline{28.00} & 63.66 \\
Uni-OPD    & \underline{86.59} & 77.78 & \underline{28.00} & \underline{64.12} \\
ExOPD      & 85.37 & 76.98 & \textbf{29.14} & 63.83 \\
\textbf{RA-OPD}
& \textbf{87.80} & \underline{78.57} & \textbf{29.14}
& \textbf{65.17} \\
\bottomrule
\end{tabular*}
\setcounter{table}{2}
\caption{Code generation results (\%) with Qwen3-8B-Base-GRPO-Code as the teacher and Qwen3-4B-Base as the student.  HE and LCB abbreviate HumanEval and LiveCodeBench, respectively.  The best and second-best results are shown in bold and underlined, respectively.}
\label{tab:code}
\end{table}
}}}

\begin{findingbox}
\textbf{Finding 2.} RA-OPD maintains its effectiveness consistently across the DeepSeek-R1 family.
\end{findingbox}

RA-OPD achieves the strongest overall performance on both avg@$k$ and pass@$k$ with Skywork-OR1-Math-7B as the teacher and DeepSeek-R1-Distill-Qwen-7B as the student. Table~\ref{tab:deepseek-results} reports the full results. RA-OPD achieves an average avg@$k$ of 69.34, surpassing standard OPD by 4.91 points and the strongest competing baseline, Uni-OPD, by 2.79 points. On AIME24, RA-OPD reaches 65.62, improving over OPD by 8.43 points. On AIME25, it reaches 55.31, an improvement of 9.16 points. On AIME26, it achieves 64.17, exceeding OPD by 8.86 points. The gains also extend to the remaining benchmarks. RA-OPD reaches 88.70 on AMC, improving OPD by 2.10 points, and 96.53 on MATH, a gain of 1.38 points. On Minerva Math, it achieves 41.13, improving OPD by 2.62 points. On OlympiadBench, it reaches 73.93, exceeding OPD by 1.84 points. RA-OPD also achieves the highest average pass@$k$ of 83.48, surpassing standard OPD, the strongest competing baseline on this metric. For average pass@$k$, RA-OPD exceeds Uni-OPD by 3.53 points. It also surpasses ExOPD by 4.52 points. These results confirm that RA-OPD maintains its effectiveness consistently across the DeepSeek-R1 family.

\begin{findingbox}
\textbf{Finding 3.} RA-OPD maintains training efficiency comparable to that of standard OPD and requires less than one third of Uni-OPD's training time.
\end{findingbox}

RA-OPD maintains training efficiency comparable to standard OPD while using less than one third of Uni-OPD's total training time. For Uni-OPD, we sample four responses per prompt following its default setting. Figure~\ref{fig:training-time} compares the end-to-end time required by each optimization step during training. OPD, ExOPD, and RA-OPD remain in a similar timing regime, whereas Uni-OPD is consistently much slower because its group-based training pipeline incurs substantially greater rollout and optimization cost. Summing the per-step measurements over the plotted training interval gives total times of 4.38 hours for OPD, 4.77 hours for ExOPD, 14.41 hours for Uni-OPD, and 4.48 hours for RA-OPD. RA-OPD therefore adds only 0.10 hours relative to standard OPD and is 0.29 hours faster than ExOPD, while reducing the total training time by 9.93 hours relative to Uni-OPD. Equivalently, Uni-OPD requires approximately $3.22\times$ as much time as RA-OPD. These results show that reward-alignment filtering preserves the practical efficiency of standard OPD because it reuses the existing rollout and teacher signals instead of introducing additional rollout groups.

\begin{findingbox}
\textbf{Finding 4.} Reward-misaligned trajectories remain prevalent throughout training in both model families.
\end{findingbox}

RA-OPD identifies a substantial fraction of reward-misaligned trajectories throughout training in both model settings. Figure~\ref{fig:drop-fractions} summarizes these return and filtering dynamics. At each fixed prefix, because tokens are sampled from the student policy, the expected token-level teacher guidance equals the negative reverse KL and is therefore non-positive. Consistent with this property, mean $G_i$ remains below zero throughout training, while $P(G_i<0)$ averages 94.25\% in the Qwen3 setting and 92.46\% in the DeepSeek-R1 setting. The overall dropped fraction averages 48.63\% and 68.23\%, respectively, showing that reward-misaligned trajectories are prevalent in both model families. At the beginning of training, nearly all correct trajectories are filtered, whereas nearly all incorrect trajectories are kept. As training proceeds, more correct trajectories are kept and more incorrect trajectories are filtered. We conjecture that this change reflects the improving student policy. The student gradually produces correct responses to some difficult questions, and the teacher may assign these trajectories higher probabilities than the student does. The resulting positive return is consistent with the outcome reward, so RA-OPD keeps these correct trajectories. Meanwhile, the teacher may also assign some incorrect trajectories higher probabilities than the student does. The resulting positive return is inconsistent with the outcome reward, so RA-OPD filters these incorrect trajectories. These dynamics validate RA-OPD's motivation.

\subsection{Results on Code}
\label{subsec:code-results}

\begin{findingbox}
\textbf{Finding 5.} RA-OPD achieves the strongest average performance among the tested OPD methods on code generation benchmarks.
\end{findingbox}

RA-OPD achieves the best result on code benchmarks as shown in Table~\ref{tab:code}.  Its average of 58.97 exceeds standard OPD by 3.93 points and ExOPD, the strongest competing baseline, by 2.87 points.  Specifically, it reaches 78.81 on HumanEval$+$, 70.97 on MBPP$+$, and 27.14 on LiveCodeBench v6, yielding gains of 6.71, 2.25, and 2.85 points over OPD, respectively.  The improvement therefore covers both unit-test-based synthesis benchmarks and the more recent contest problems in LiveCodeBench.  Mean pass@4 exhibits the same aggregate trend: RA-OPD reaches 65.17, 1.51 points above OPD and 1.05 points above Uni-OPD, the strongest competing baseline on this metric.

\subsection{Ablation Study}

Our ablations show that masking either correct-negative or incorrect-positive conflicts improves standard OPD, while masking both yields the strongest performance. Training only on the two conflict types still improves over the untrained base model but underperforms both standard OPD and RA-OPD, indicating that reward-misaligned trajectories contain useful token-level information but have a lower signal-to-noise ratio. RA-OPD also closely preserves the actor-entropy profile of standard OPD across both model families, showing that trajectory filtering does not broadly alter exploration dynamics. Details are provided in Appendix~\ref{sec:additional-experiments}.

\section{Conclusion}
\label{sec:conclusion}

In this work, we identified reward-misaligned teacher guidance as a key reliability problem in OPD. On student-generated prefixes, such guidance may discourage correct trajectories or encourage incorrect ones, misleading optimization and degrading model performance. We therefore proposed Reward-Aligned On-Policy Distillation (RA-OPD), which filters trajectories whose trajectory-level distillation returns are inconsistent with their outcome rewards. By reusing existing student trajectories and teacher probabilities, RA-OPD requires no additional student rollouts or teacher evaluations, and its absolute criterion remains well defined with one trajectory per prompt and does not require reward diversity. Experiments using Qwen3 and DeepSeek-R1 models show that reward-misaligned trajectories remain prevalent throughout training and that, across seven math and three code benchmarks, RA-OPD achieves the strongest average performance while maintaining efficiency comparable to standard OPD.

\bibliography{references}

@inproceedings{agarwal2024onpolicy,
  title={On-policy distillation of language models: Learning from self-generated mistakes},
  author={Agarwal, Rishabh and Vieillard, Nino and Zhou, Yongchao and Stanczyk, Piotr and Ramos Garea, Sabela and Geist, Matthieu and Bachem, Olivier},
  booktitle={Proceedings of the Twelfth International Conference on Learning Representations},
  volume={2024},
  pages={21246--21263},
  year={2024}
}

@misc{armandpour2026unmasking,
  title={Unmasking on-policy distillation: Where it helps, where it hurts, and why},
  author={Armandpour, Mohammadreza and Ilhan, Fatih and Harrison, David and Jaiswal, Ajay and Hoang, Duc NM and Faghri, Fartash and Zhang, Yizhe and Cho, Minsik and Farajtabar, Mehrdad},
  archivePrefix={arXiv},
  eprint={2605.10889},
  year={2026}
}

@inproceedings{bengio2015scheduled,
  title={Scheduled sampling for sequence prediction with recurrent neural networks},
  author={Bengio, Samy and Vinyals, Oriol and Jaitly, Navdeep and Shazeer, Noam},
  booktitle={Proceedings of the Twenty-eighth Annual Conference on Neural Information Processing Systems},
  volume={28},
  year={2015}
}

@inproceedings{gu2024minillm,
  title={Minillm: Knowledge distillation of large language models},
  author={Gu, Yuxian and Dong, Li and Wei, Furu and Huang, Minlie},
  booktitle={Proceedings of the Twelfth International Conference on Learning Representations},
  year={2024}
}

@article{guo2025deepseekr1,
  title={DeepSeek-R1 incentivizes reasoning in LLMs through reinforcement learning},
  author={Guo, Daya and Yang, Dejian and Zhang, Haowei and Song, Junxiao and Wang, Peiyi and Zhu, Qihao and Xu, Runxin and Zhang, Ruoyu and Ma, Shirong and Bi, Xiao and others},
  journal={Nature},
  volume={645},
  number={8081},
  pages={633--638},
  year={2025},
  publisher={Nature Publishing Group UK London}
}

@misc{gudibande2023false,
  title={The false promise of imitating proprietary llms},
  author={Gudibande, Arnav and Wallace, Eric and Snell, Charlie and Geng, Xinyang and Liu, Hao and Abbeel, Pieter and Levine, Sergey and Song, Dawn},
  archivePrefix={arXiv},
  eprint={2305.15717},
  year={2023}
}

@misc{hinton2015distilling,
  title={Distilling the knowledge in a neural network},
  author={Hinton, Geoffrey and Vinyals, Oriol and Dean, Jeff},
  eprint={1503.02531},
  archivePrefix={arXiv},
  year={2015}
}

@misc{hou2026uniopd,
  title={Uni-opd: Unifying on-policy distillation with a dual-perspective recipe},
  author={Hou, Wenjin and Peng, Shangpin and Wang, Weinong and Ruan, Zheng and Zhang, Yue and Zhou, Zhenglin and Gao, Mingqi and Chen, Yifei and Wang, Kaiqi and Yang, Hongming and others},
  eprint={2605.03677},
  archivePrefix={arXiv},
  year={2026}
}

@inproceedings{jang2026stable,
    title = "Stable On-Policy Distillation through Adaptive Target Reformulation",
    author = "Jang, Ijun  and
      Yeom, Jewon  and
      Yeo, Juan  and
      Lim, Hyunggyu  and
      Kim, Taesup",
    editor = "Liakata, Maria  and
      Moreira, Viviane P.  and
      Zhang, Jiajun  and
      Jurgens, David",
    booktitle = "Findings of the Association for Computational Linguistics: ACL 2026",
    month = jul,
    year = "2026",
    address = "San Diego, California, United States",
    publisher = "Association for Computational Linguistics",
}

@misc{jiang2026trajectory,
    title={Trajectory-Refined Distillation}, 
    author={Li Jiang and Haoran Xu and Yichuan Ding and Amy Zhang},
    year={2026},
    eprint={2606.08432},
    archivePrefix={arXiv},
    primaryClass={cs.AI},
}

@inproceedings{jiao2020tinybert,
    title = "{T}iny{BERT}: Distilling {BERT} for Natural Language Understanding",
    author = "Jiao, Xiaoqi  and
      Yin, Yichun  and
      Shang, Lifeng  and
      Jiang, Xin  and
      Chen, Xiao  and
      Li, Linlin  and
      Wang, Fang  and
      Liu, Qun",
    editor = "Cohn, Trevor  and
      He, Yulan  and
      Liu, Yang",
    booktitle = "Findings of the Association for Computational Linguistics: EMNLP 2020",
    month = nov,
    year = "2020",
    address = "Online",
    publisher = "Association for Computational Linguistics",
}

@misc{jin2026eopd,
    title={Entropy-Aware On-Policy Distillation of Language Models}, 
    author={Woogyeol Jin and Taywon Min and Yongjin Yang and Dennis Wei and Yi Zhou and Swanand Ravindra Kadhe and Nathalie Baracaldo and Kimin Lee},
    year={2026},
    eprint={2603.07079},
    archivePrefix={arXiv},
    primaryClass={cs.LG},
    url={https://arxiv.org/abs/2603.07079}, 
}

@inproceedings{kim2016sequence,
    title = "Sequence-Level Knowledge Distillation",
    author = "Kim, Yoon  and
      Rush, Alexander M.",
    editor = "Su, Jian  and
      Duh, Kevin  and
      Carreras, Xavier",
    booktitle = "Proceedings of the 2016 Conference on Empirical Methods in Natural Language Processing",
    month = nov,
    year = "2016",
    address = "Austin, Texas",
    publisher = "Association for Computational Linguistics",
}

@misc{li2026sample,
    title={Unifying Group-Relative and Self-Distillation Policy Optimization via Sample Routing}, 
    author={Gengsheng Li and Tianyu Yang and Junfeng Fang and Mingyang Song and Mao Zheng and Haiyun Guo and Dan Zhang and Jinqiao Wang and Tat-Seng Chua},
    year={2026},
    eprint={2604.02288},
    archivePrefix={arXiv},
    primaryClass={cs.LG},
    url={https://arxiv.org/abs/2604.02288}, 
}

@misc{li2026rethinking,
    title={Rethinking On-Policy Distillation of Large Language Models: Phenomenology, Mechanism, and Recipe}, 
    author={Yaxuan Li and Yuxin Zuo and Bingxiang He and Jinqian Zhang and Chaojun Xiao and Cheng Qian and Tianyu Yu and Gao, Huan-ang and Wenkai Yang and Zhiyuan Liu and Ning Ding},
    year={2026},
    eprint={2604.13016},
    archivePrefix={arXiv},
    primaryClass={cs.LG},
}

@misc{liu2026teacher,
    title={Your Teacher Can't Help You Here: Combating Supervision Fidelity Decay in On-Policy Distillation}, 
    author={Yanjiang Liu and Jie Lou and Xinyan Guan and Yuqiu Ji and Hongyu Lin and Ben He and Xianpei Han and Le Sun and Xing Yu and Yaojie Lu},
    year={2026},
    archivePrefix={arXiv},
    eprint={2605.30833},
    primaryClass={cs.CL},
}

@article{lu2025onpolicydistillation,
  author = {Kevin Lu and Thinking Machines Lab},
  title = {On-Policy Distillation},
  journal = {Thinking Machines Lab: Connectionism},
  year = {2025},
  note = {https://thinkingmachines.ai/blog/on-policy-distillation},
  doi = {10.64434/tml.20251026},
}

@misc{luo2026demystifying,
    title={Demystifying OPD: Length Inflation and Stabilization Strategies for Large Language Models}, 
    author={Feng Luo and Yu-Neng Chuang and Guanchu Wang and Zicheng Xu and Xiaotian Han and Tianyi Zhang and Vladimir Braverman},
    year={2026},
    eprint={2604.08527},
    archivePrefix={arXiv},
    primaryClass={cs.CL},
}

@misc{oh2026vopd,
    title={KL for a KL: On-Policy Distillation with Control Variate Baseline}, 
    author={Minjae Oh and Sangjun Song and Gyubin Choi and Yunho Choi and Yohan Jo},
    year={2026},
    eprint={2605.07865},
    archivePrefix={arXiv},
    primaryClass={cs.LG},
}

@inproceedings{ross2011dagger,
  title={A reduction of imitation learning and structured prediction to no-regret online learning},
  author={Ross, St{\'e}phane and Gordon, Geoffrey and Bagnell, Drew},
  booktitle={Proceedings of the Fourteenth International Conference on Artificial Intelligence and Statistics},
  pages={627--635},
  year={2011},
  organization={JMLR Workshop and Conference Proceedings}
}

@misc{sanh2019distilbert,
    title={DistilBERT, a distilled version of BERT: smaller, faster, cheaper and lighter}, 
    author={Victor Sanh and Lysandre Debut and Julien Chaumond and Thomas Wolf},
    year={2020},
    eprint={1910.01108},
    archivePrefix={arXiv},
    primaryClass={cs.CL},
    url={https://arxiv.org/abs/1910.01108}, 
}

@misc{schulman2017ppo,
  title={Proximal policy optimization algorithms},
  author={Schulman, John and Wolski, Filip and Dhariwal, Prafulla and Radford, Alec and Klimov, Oleg},
  archivePrefix={arXiv},
  eprint={1707.06347},
  year={2017}
}

@misc{shao2024deepseekmath,
    title={DeepSeekMath: Pushing the Limits of Mathematical Reasoning in Open Language Models}, 
    author={Zhihong Shao and Peiyi Wang and Qihao Zhu and Runxin Xu and Junxiao Song and Xiao Bi and Haowei Zhang and Mingchuan Zhang and Y. K. Li and Y. Wu and Daya Guo},
    year={2024},
    eprint={2402.03300},
    archivePrefix={arXiv},
    primaryClass={cs.CL},
}

@misc{song2026survey,
    title={A Survey of On-Policy Distillation for Large Language Models}, 
    author={Mingyang Song and Mao Zheng},
    year={2026},
    eprint={2604.00626},
    archivePrefix={arXiv},
    primaryClass={cs.LG},
}

@misc{xiao2026mimo,
  title={Mimo-v2-flash technical report},
  author={Xiao, Bangjun and Xia, Bingquan and Yang, Bo and Gao, Bofei and Shen, Bowen and Zhang, Chen and He, Chenhong and Lou, Chiheng and Luo, Fuli and Wang, Gang and others},
  eprint={2601.02780},
  archivePrefix={arXiv},
  year={2026}
}

@misc{xu2026sgopd,
    title={SG-OPD: Sign-Gated On-Policy Distillation via Sign-Consistency Gating and Phased Teacher Sampling}, 
    author={Haoran Xu and Hongyu Wang and Yifei Gao and Jiaze Li and Xiaofeng Zhang and Xiaosong Yuan},
    year={2026},
    eprint={2606.09304},
    archivePrefix={arXiv},
    primaryClass={cs.CL},
}

@misc{yang2025qwen3,
  title={Qwen3 technical report},
  author={Yang, An and Li, Anfeng and Yang, Baosong and Zhang, Beichen and Hui, Binyuan and Zheng, Bo and Yu, Bowen and Gao, Chang and Huang, Chengen and Lv, Chenxu and others},
  archivePrefix={arXiv},
  eprint={2505.09388},
  year={2025}
}

@misc{yang2026beyond,
    title={Learning beyond Teacher: Generalized On-Policy Distillation with Reward Extrapolation}, 
    author={Wenkai Yang and Weijie Liu and Ruobing Xie and Kai Yang and Saiyong Yang and Yankai Lin},
    year={2026},
    eprint={2602.12125},
    archivePrefix={arXiv},
    primaryClass={cs.LG},
}

@misc{yang2026pruneopd,
    title={Prune-OPD: Efficient and Reliable On-Policy Distillation for Long-Horizon Reasoning}, 
    author={Zhicheng Yang and Zhijiang Guo and Yifan Song and Minrui Xu and Yongxin Wang and Yiwei Wang and Xiaodan Liang and Jing Tang},
    year={2026},
    eprint={2605.07804},
    archivePrefix={arXiv},
    primaryClass={cs.LG},
}

@misc{zeng2026glm5,
  title={Glm-5: from vibe coding to agentic engineering},
  author={Zeng, Aohan and Lv, Xin and Hou, Zhenyu and Du, Zhengxiao and Zheng, Qinkai and Chen, Bin and Yin, Da and Ge, Chendi and Huang, Chenghua and Xie, Chengxing and others},
  archivePrefix={arXiv},
  eprint={2602.15763},
  year={2026}
}

@misc{zhou2026less,
    title={Less is More: Early Stopping Rollout for On-Policy Distillation}, 
    author={Zhou Ziheng and Jiaqi Li and Huacong Tang and Ying Nian Wu and Demetri Terzopoulos},
    year={2026},
    eprint={2605.27028},
    archivePrefix={arXiv},
    primaryClass={cs.LG},
}

@misc{zhu2026many,
    title={The Many Faces of On-Policy Distillation: Pitfalls, Mechanisms, and Fixes}, 
    author={Siqi Zhu and Xuyan Ye and Hongyu Lu and Weiye Shi and Ge Liu},
    year={2026},
    eprint={2605.11182},
    archivePrefix={arXiv},
    primaryClass={cs.AI},
}

@misc{he2025skyworkor1,
    title={Skywork Open Reasoner 1 Technical Report}, 
    author={Jujie He and Jiacai Liu and Chris Yuhao Liu and Rui Yan and Chaojie Wang and Peng Cheng and Xiaoyu Zhang and Fuxiang Zhang and Jiacheng Xu and Wei Shen and Siyuan Li and Liang Zeng and Tianwen Wei and Cheng Cheng and Bo An and Yang Liu and Yahui Zhou},
    year={2025},
    eprint={2505.22312},
    archivePrefix={arXiv},
    primaryClass={cs.LG},
}

@inproceedings{hendrycks2021math,
 author = {Hendrycks, Dan and Burns, Collin and Kadavath, Saurav and Arora, Akul and Basart, Steven and Tang, Eric and Song, Dawn and Steinhardt, Jacob},
 booktitle = {Proceedings of the Neural Information Processing Systems Track on Datasets and Benchmarks},
 editor = {J. Vanschoren and S. Yeung},
 pages = {},
 title = {Measuring Mathematical Problem Solving With the MATH Dataset},
 volume = {1},
 year = {2021}
}

@inproceedings{lewkowycz2022minerva,
  title={Solving quantitative reasoning problems with language models},
  author={Lewkowycz, Aitor and Andreassen, Anders and Dohan, David and Dyer, Ethan and Michalewski, Henryk and Ramasesh, Vinay and Slone, Ambrose and Anil, Cem and Schlag, Imanol and Gutman-Solo, Theo and others},
  booktitle={Proceedings of the Thirty-fifth Annual Conference on Neural Information Processing Systems},
  volume={35},
  pages={3843--3857},
  year={2022}
}

@inproceedings{he2024olympiadbench,
  title={Olympiadbench: A challenging benchmark for promoting agi with olympiad-level bilingual multimodal scientific problems},
  author={He, Chaoqun and Luo, Renjie and Bai, Yuzhuo and Hu, Shengding and Thai, Zhen and Shen, Junhao and Hu, Jinyi and Han, Xu and Huang, Yujie and Zhang, Yuxiang and others},
  booktitle={Proceedings of the Sixty-second Annual Meeting of the Association for Computational Linguistics},
  pages={3828--3850},
  year={2024}
}

@inproceedings{liu2023evalplus,
  title={Is your code generated by chatgpt really correct? rigorous evaluation of large language models for code generation},
  author={Liu, Jiawei and Xia, Chunqiu Steven and Wang, Yuyao and Zhang, Lingming},
  booktitle={Proceedings of the Thirty-sixth Annual Conference on Neural Information Processing Systems},
  volume={36},
  pages={21558--21572},
  year={2023}
}

@misc{jain2024livecodebench,
  title={Livecodebench: Holistic and contamination free evaluation of large language models for code},
  author={Jain, Naman and Han, King and Gu, Alex and Li, Wen-Ding and Yan, Fanjia and Zhang, Tianjun and Wang, Sida and Solar-Lezama, Armando and Sen, Koushik and Stoica, Ion},
  archivePrefix={arXiv},
  eprint={2403.07974},
  year={2024}
}

@misc{chen2021evaluating,
  title={Evaluating large language models trained on code},
  author={Chen, Mark and Tworek, Jerry and Jun, Heewoo and Yuan, Qiming and Pinto, Henrique Ponde De Oliveira and Kaplan, Jared and Edwards, Harri and Burda, Yuri and Joseph, Nicholas and Brockman, Greg and others},
  archivePrefix={arXiv},
  eprint={2107.03374},
  year={2021}
}

@misc{austin2021program,
  title={Program synthesis with large language models},
  author={Austin, Jacob and Odena, Augustus and Nye, Maxwell and Bosma, Maarten and Michalewski, Henryk and Dohan, David and Jiang, Ellen and Cai, Carrie and Terry, Michael and Le, Quoc and others},
  archivePrefix={arXiv},
  eprint={2108.07732},
  year={2021}
}

\clearpage
\appendix
\onecolumn
\raggedbottom
\section{Related Work}
\enlargethispage{2\baselineskip}

\subsection{On-Policy Distillation}

Knowledge distillation transfers the knowledge and capabilities of a teacher model into a student model by training the student to match the teacher distribution~\cite{hinton2015distilling,sanh2019distilbert,jiao2020tinybert}. In conventional off-policy distillation for autoregressive language models, training trajectories are drawn from a fixed corpus, typically consisting of original training data or teacher-generated trajectories~\cite{kim2016sequence,gudibande2023false,song2026survey}. At each position, the teacher and student condition on prefixes from this fixed corpus. During inference, the student instead generates autoregressively from its own previous outputs. A deviation from the fixed training trajectories changes the prefixes on which all subsequent predictions are conditioned, creating a training--inference prefix mismatch that can accumulate along the trajectory~\cite{ross2011dagger,bengio2015scheduled,song2026survey}.

On-policy distillation replaces the static corpus with trajectories sampled from the current student policy. The teacher evaluates the student-generated prefixes that the student actually visits, so the training distribution evolves with the student policy and more closely matches its inference-time behavior~\cite{agarwal2024onpolicy,gu2024minillm,lu2025onpolicydistillation,song2026survey}. Under the reverse KL objective considered in this work, the feedback at each position is computed from the teacher and student next-token probabilities. This formulation alleviates the training--inference prefix mismatch and has been adopted in the post-training pipelines of recent models such as Qwen3, MiMo, and GLM-5~\cite{yang2025qwen3,xiao2026mimo,zeng2026glm5,li2026rethinking}.

Existing OPD implementations differ in how much of the teacher distribution they use at each position~\cite{jang2026stable,jin2026eopd,oh2026vopd,zhu2026many}. Sampled-token OPD uses only the teacher probability of the token sampled by the student, Top-$k$ OPD uses a subset of candidate tokens, and full-vocabulary OPD uses the entire vocabulary. For a sampled token $o_t$, the feedback is the log-probability difference $\log \pi_T(o_t \mid q,o_{<t})-\log \pi_\theta(o_t \mid q,o_{<t})$. This lightweight implementation is widely used~\cite{lu2025onpolicydistillation,oh2026vopd,song2026survey}; we adopt it without changing the standard OPD objective.

\subsection{Reliable Teacher Guidance in OPD}

Student-generated prefixes may contain reasoning errors or fall outside the teacher's training distribution, making teacher guidance unreliable and potentially degrading or even destabilizing OPD~\cite{armandpour2026unmasking,li2026sample,li2026rethinking,luo2026demystifying,song2026survey,zhu2026many}. Existing analyses connect guidance reliability to final model performance but do not provide a generally effective solution~\cite{armandpour2026unmasking,li2026rethinking,song2026survey,zhu2026many}.

Several methods assess and modify teacher guidance locally~\cite{jang2026stable,jin2026eopd,oh2026vopd,xu2026sgopd}. Early Stopping Rollout (ESR) restricts each student rollout to its first $N$ response tokens and computes the distillation loss only over this fixed early window~\cite{zhou2026less}. Lookahead Group Reward (LGR) evaluates the student's top-$k$ candidate tokens using the teacher confidence they induce at the next step and assigns a group-normalized reward~\cite{liu2026teacher}. Prune-OPD monitors local student--teacher compatibility, reweights subsequent OPD rewards after prefix-drift events, and truncates the response when cumulative drift exceeds a budget~\cite{yang2026pruneopd}. Together, these methods restrict, reweight, or truncate local teacher guidance, making them complementary to RA-OPD's trajectory-level consistency check. However, their local criteria alone do not determine whether the overall update induced by a complete response is consistent with its final outcome accuracy. Misleading guidance may therefore still move the student toward an incorrect response or discourage it from moving toward a correct one~\cite{jiang2026trajectory,li2026sample,xu2026sgopd}.

Uni-OPD aggregates the teacher guidance along a response into a trajectory-level distillation return and applies outcome-guided margin calibration within each prompt's rollout group~\cite{hou2026uniopd}. By enforcing that correct trajectories rank above incorrect ones, its criterion relies on relative comparisons among multiple trajectories rather than determining each trajectory's absolute update direction. If both returns are negative, a correct trajectory can rank higher while still being discouraged; if both are positive, the incorrect trajectory is still reinforced.

Reward-Aligned On-Policy Distillation (RA-OPD) instead checks whether each trajectory's distillation return is consistent with its final outcome accuracy. It keeps updates that move the student toward correct responses or away from incorrect ones and filters out the two conflicting cases. This group-free criterion determines each trajectory's absolute reliability without requiring multiple trajectories for the same prompt. RA-OPD preserves the standard OPD objective on the kept samples and requires neither Top-$k$ nor full-vocabulary teacher probabilities.

\clearpage
\section{Additional Experiments}
\label{sec:additional-experiments}

\begin{figure}[H]
\centering
\includegraphics[width=0.55\textwidth]{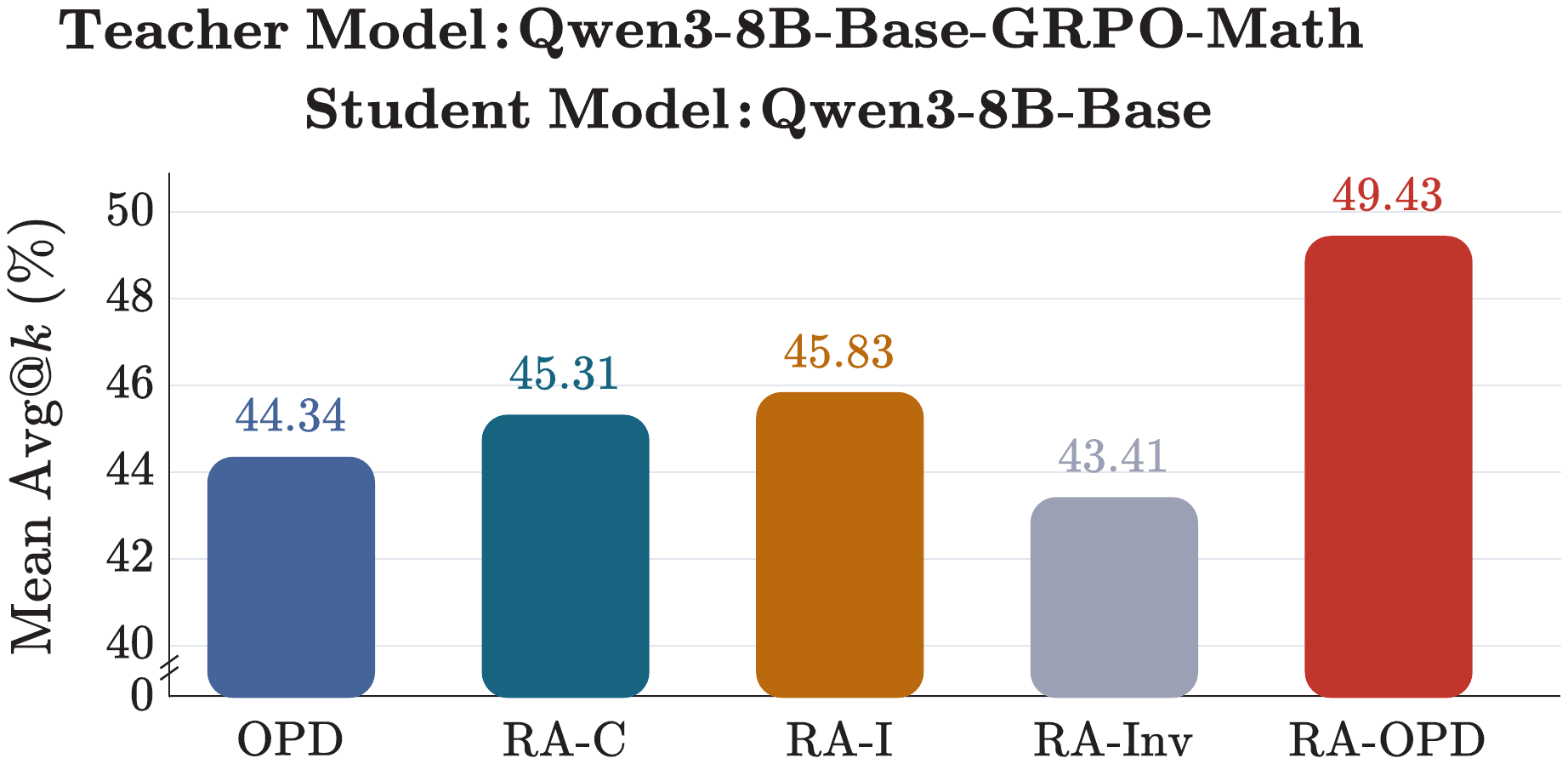}
\caption{Mean avg@$k$ across the seven math benchmarks for the ablations in Table~\ref{tab:mask-ablation}.  RA-C and RA-I mask only correct-negative and incorrect-positive conflicts, respectively, whereas RA-Inv retains only these two conflict types.}
\label{fig:inverse-mask}
\end{figure}

\paragraph{Ablation notation.} RA-C denotes the variant that masks only correct-negative ($R{=}1,G{<}0$) conflicts, with C indicating correct-negative.  Similarly, RA-I denotes the variant that masks only incorrect-positive ($R{=}0,G{>}0$) conflicts, with I indicating incorrect-positive.  RA-Inv denotes the inverse variant, which retains only the two conflict types removed by RA-OPD.  These abbreviations are adopted consistently in Table~\ref{tab:mask-ablation} and Figure~\ref{fig:inverse-mask}.

\begin{table}[H]
\centering
\begin{tabular}{lccc}
\toprule
\multicolumn{1}{c}{\textbf{Method}} & \textbf{Rule}
& \textbf{avg@$k$} & \textbf{pass@$k$} \\
\midrule
OPD & All & 44.34 & 57.66 \\
RA-C & C-negative & 45.31 & 60.63 \\
RA-I & I-positive & 45.83 & 59.75 \\
RA-Inv & Conflicts & 43.41 & 56.80 \\
\textbf{RA-OPD} & Both & \textbf{49.43} & \textbf{61.56} \\
\bottomrule
\end{tabular}
\setcounter{table}{3}
\caption{Reward-alignment ablations on Qwen3-8B-Base, averaged across the seven math benchmarks (\%).  RA-C and RA-I mask only correct-negative ($R{=}1,G{<}0$) and incorrect-positive ($R{=}0,G{>}0$) conflicts, respectively, whereas RA-Inv retains only these two conflict types.}
\label{tab:mask-ablation}
\end{table}

\begin{findingbox}
\textbf{Finding 6.} Misaligned trajectories contain useful learning signal, but using them alone is worse than standard OPD, indicating a lower signal-to-noise ratio.
\end{findingbox}

Figure~\ref{fig:inverse-mask} evaluates RA-Inv by training the student using only the two types of trajectories that RA-OPD would remove.  RA-Inv reaches 43.41 average avg@$k$, substantially above the untrained base model (26.30).  Thus, misaligned trajectories are not devoid of useful token-level information.  Nevertheless, RA-Inv is 0.93 points below standard OPD and 6.02 points below RA-OPD.  Average pass@$k$ shows the same ordering (56.80 versus 57.66 and 61.56).  The result is consistent with these trajectories mixing useful local teacher information with noisy or outcome-conflicting trajectory-level updates; retaining only them leaves a weaker training signal than either OPD or reward-aligned filtering.

\begin{findingbox}
\textbf{Finding 7.} Masking either type of reward-alignment conflict helps, while masking both is necessary for the full improvement of RA-OPD.
\end{findingbox}

Table~\ref{tab:mask-ablation} separates the two failure modes in the reward-alignment criterion.  RA-C, which removes only correct-negative trajectories, improves average avg@$k$ from 44.34 to 45.31.  RA-I, which removes only incorrect-positive trajectories, raises it to 45.83.  Both one-sided variants therefore improve over OPD, but remain 3.60--4.12 points below the complete RA-OPD result of 49.43.  The average pass@$k$ results are consistent: RA-C and RA-I reach 60.63 and 59.75, respectively, while removing both conflicts reaches 61.56.  These complementary gains empirically support both halves of the reward-alignment rule.

\begin{figure}[H]
\centering
\hspace*{0.01\textwidth}%
\includegraphics[width=0.47\textwidth]{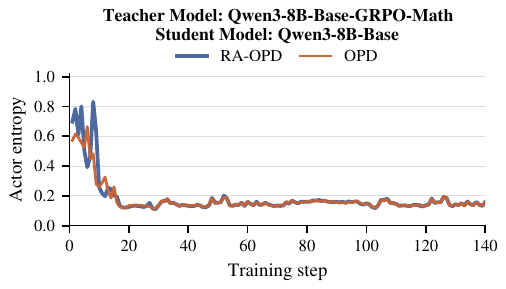}\hfill
\includegraphics[width=0.47\textwidth]{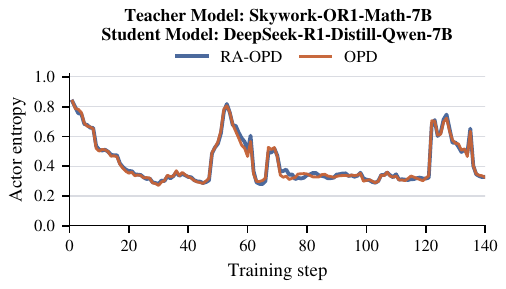}%
\hspace*{0.01\textwidth}
\caption{Actor entropy for RA-OPD and standard OPD during training.  The left and right panels show Qwen3-8B-Base and DeepSeek-R1-Distill-Qwen-7B, respectively.}
\label{fig:actor-entropy}
\end{figure}

\begin{findingbox}
\textbf{Finding 8.} RA-OPD preserves the actor-entropy profile of standard OPD across model families.
\end{findingbox}

Despite filtering a large fraction of trajectories, RA-OPD closely tracks the entropy profile of standard OPD in both the Qwen3 and DeepSeek-series settings (Figure~\ref{fig:actor-entropy}).  During training, mean actor entropy is 0.183 for RA-OPD and 0.177 for OPD on Qwen3-8B-Base, and 0.424 and 0.422, respectively, on DeepSeek-R1-Distill-Qwen-7B.  The Qwen3 runs differ mainly during the initial stage and become nearly indistinguishable thereafter, while the DeepSeek-family curves remain closely overlapped throughout training.  Together with the high conflict rates in Figure~\ref{fig:drop-fractions}, this shows that RA-OPD changes which trajectory-level updates are trusted without inducing a broad shift in the actor's exploration dynamics.

\clearpage
\section{Training Parameters}
\label{app:training-parameters}

This section separates the settings used for model training from those used only for final evaluation.

\subsection{Training-Time Parameters}
\label{app:training-time-parameters}

For standard OPD and RA-OPD, the student samples one on-policy response per training prompt through vLLM, after which the teacher scores the sampled trajectory and the student is updated with the sampled-token k1 OPD objective. We train every student for two epochs. For baseline-specific experimental settings prescribed by ExOPD and Uni-OPD, including ExOPD's reward-scaling factor $\lambda=1.25$ and Uni-OPD's use of four responses per prompt, we use the best-performing hyperparameters reported in their respective papers; all remaining optimization and rollout hyperparameters are held fixed across methods. ExOPD uses the student base model as its reference policy. Relative to standard OPD, RA-OPD's only method-specific change is reward-alignment filtering: RA-OPD masks misaligned trajectories, whereas standard OPD disables this option. Table~\ref{tab:training-parameters} lists the shared training-time configuration and the baseline-specific exceptions.

\begin{table}[H]
\centering
\small
\setlength{\tabcolsep}{4pt}
\renewcommand{\arraystretch}{1.0}
\begin{tabular}{
  >{\raggedright\arraybackslash}p{0.32\textwidth}
  >{\raggedright\arraybackslash}p{0.24\textwidth}
  >{\raggedright\arraybackslash}p{0.36\textwidth}}
\toprule
\textbf{Parameter} & \textbf{Value} & \textbf{Description} \\
\midrule
data\_\allowbreak train\_\allowbreak files
  & DAPO-Math-17K (math); Eurus-2-RL-Data-Code-25K (code)
  & Task-specific training set. \\
data\_\allowbreak train\_\allowbreak batch\_\allowbreak size
  & 256
  & Global prompt batch size and maximum rollout/teacher batch. \\
trainer.total\_\allowbreak epochs
  & 2
  & Number of passes over the OPD training set. \\
actor.optim.lr
  & $1\times10^{-6}$
  & Student-model learning rate. \\
data\_\allowbreak max\_\allowbreak prompt\_\allowbreak length
  & 2048
  & Maximum prompt length in tokens. \\
data\_\allowbreak max\_\allowbreak response\_\allowbreak length
  & 8192
  & Maximum training response length (16384 for DeepSeek). \\
rollout.val\_\allowbreak response\_\allowbreak length
  & 16384
  & Maximum response length used during validation. \\

rollout.name
  & vllm
  & Rollout backend. \\
rollout.tensor\_\allowbreak model\_\allowbreak parallel\_\allowbreak size
  & 1
  & Student rollout tensor-parallel size. \\
rollout.gpu\_\allowbreak memory\_\allowbreak utilization
  & 0.6
  & vLLM GPU-memory utilization target. \\
rollout.n
  & 1 (4 for Uni-OPD)
  & One response per prompt by default; Uni-OPD uses four following its reported setting. \\
rollout.max\_\allowbreak model\_\allowbreak len
  & 18433
  & $2048+\max(8192,16384)+1$. \\
rollout.max\_\allowbreak num\_\allowbreak batched\_\allowbreak tokens
  & 18433
  & Matches the actor maximum model length. \\
rollout.max\_\allowbreak num\_\allowbreak seqs
  & 256
  & Matches the training batch size. \\
rollout.enforce\_\allowbreak eager
  & True
  & Uses eager execution for vLLM. \\
rollout.calculate\_\allowbreak log\_\allowbreak probs
  & False
  & Does not request rollout-side log probabilities. \\
rollout.temperature
  & 1.0
  & Training-rollout sampling temperature. \\
rollout.top\_\allowbreak p
  & 1.0
  & Training-rollout nucleus threshold. \\
rollout.top\_\allowbreak k
  & $-1$
  & Disables top-$k$ truncation for training rollouts. \\

actor.policy\_\allowbreak loss.lambda\_\allowbreak vals
  & 1.25 (ExOPD)
  & Reward-extrapolation factor; standard OPD uses 1.0. \\
reference policy
  & Student base model (ExOPD)
  & Reference policy used to define ExOPD's extrapolation direction. \\
distillation\_\allowbreak loss.loss\_\allowbreak mode
  & k1
  & Sampled-token k1 OPD loss used in the main experiments. \\
distillation\_\allowbreak loss.use\_\allowbreak policy\_\allowbreak gradient
  & True
  & Enables the k1 policy-gradient form. \\
distillation\_\allowbreak loss.loss\_\allowbreak max\_\allowbreak clamp
  & 10.0
  & Upper clamp applied by the loss implementation. \\
distillation\_\allowbreak loss.log\_\allowbreak prob\_\allowbreak min\_\allowbreak clamp
  & $-10.0$
  & Lower clamp on log probabilities. \\
distillation\_\allowbreak loss.drop\_\allowbreak misaligned\_\allowbreak k1\_\allowbreak sequences
  & True (RA-OPD)
  & RA-OPD enables reward-alignment filtering; standard OPD sets this to False. \\
distillation\_\allowbreak loss.drop\_\allowbreak misaligned\_\allowbreak k1\_\allowbreak sequences\_\allowbreak mode
  & mask (RA-OPD)
  & Masks reward-misaligned sequences. \\
\bottomrule
\end{tabular}
\caption{Shared training parameters and baseline-specific exceptions used in the reported experiments.}
\label{tab:training-parameters}
\end{table}

\subsection{Evaluation-Time Parameters}
\label{app:evaluation-time-parameters}

For final evaluation, we freeze each trained checkpoint and generate responses without teacher inference or gradient computation. All methods use the same decoding configuration so that the comparison reflects the learned checkpoint rather than differences in test-time sampling. Mathematical answers are checked with math\_verify; code generations are evaluated using EvalPlus tests or the LiveCodeBench judge. Following the protocol in Section~\ref{sec:experiments}, we report both avg@$k$ and pass@$k$, using $k=32$ for AIME and AMC, $k=8$ for the other math benchmarks, and $k=4$ for the code benchmarks. Table~\ref{tab:evaluation-sampling} summarizes the evaluation-time settings.

\begin{table}[H]
\centering
\small
\setlength{\tabcolsep}{4pt}
\renewcommand{\arraystretch}{1.0}
\begin{tabular}{
  >{\raggedright\arraybackslash}p{0.25\textwidth}
  >{\raggedright\arraybackslash}p{0.32\textwidth}
  >{\raggedright\arraybackslash}p{0.35\textwidth}}
\toprule
\textbf{Parameter} & \textbf{Math} & \textbf{Code} \\
\midrule
Benchmarks
  & AIME24/25/26, AMC, MATH, Minerva Math, OlympiadBench
  & HumanEval$+$, MBPP$+$, LiveCodeBench v6 \\
Backend
  & vLLM
  & vLLM with EvalPlus or LiveCodeBench \\
Maximum prompt length
  & 2048
  & Benchmark default \\
Maximum response length
  & 32768
  & 32768 \\
Samples per problem ($k$)
  & 32 (AIME/AMC); 8 (others)
  & 4 \\
Temperature
  & 0.7
  & 0.7 \\
Top-$p$
  & 0.8
  & 0.8 \\
Top-$k$
  & 20
  & 20 \\
Correctness evaluation
  & math\_verify
  & EvalPlus tests or the LiveCodeBench judge \\
\bottomrule
\end{tabular}
\caption{Sampling parameters used for final evaluation.}
\label{tab:evaluation-sampling}
\end{table}

\clearpage
\section{Case Studies}
\label{sec:case-studies}

We show four prompts on which OPD, ExOPD, and Uni-OPD fail while RA-OPD
succeeds.  The excerpts preserve each rollout's exploration, revisions,
decisive calculation or error, and final answer.  Only repetitive passages
are marked \emph{\ldots{} continued reasoning omitted \ldots{}}.

\newtcolorbox{rolloutbox}[2][]{
  enhanced,
  colback=red!3,
  colframe=red!65!black,
  boxrule=0.45pt,
  arc=1pt,
  title={#2},
  fonttitle=\fontsize{8pt}{9pt}\selectfont\bfseries,
  fontupper=\fontsize{8pt}{9pt}\selectfont,
  coltitle=white,
  left=4pt,
  right=4pt,
  top=3pt,
  bottom=3pt,
  before skip=2pt,
  after skip=2pt,
  #1
}

\newenvironment{casepair}{
  \par\centering
  \begin{tcbraster}[
    raster width=0.99\textwidth,
    raster columns=2,
    raster equal height=rows,
    raster column skip=0.015\textwidth,
    raster left skip=0pt,
    raster right skip=0pt,
    raster before skip=2pt,
    raster after skip=2pt
  ]
}{
  \end{tcbraster}
  \par
}

\newcommand{\reasoningomitted}{%
  \par\smallskip
  \noindent\hfill\textit{\ldots{} continued reasoning omitted \ldots{}}\hfill\null
  \par\smallskip
}

% ---------------------------------------------------------------------------
% Qwen3-4B-Base sources (all line 23 of the indicated JSONL):
% OPD: eval/Qwen3-4B-Base_from_merged_hf_fF_fbtF_resp8192_k1_sp1_
%      dmk1F_dmodeza_global_step_100/aime2024/sample_036.jsonl
% ExOPD (legacy checkpoint directory named G-OPD):
% /mnt/shared-storage-gpfs2/ai4energy/menglinjian/ckpts/opd-filter/
%      eval/Qwen3-4B-Base_from_GSPO-Qwen3-8B-Base_gbs240_resp8192_
%      G-OPD_global_step_100/aime2024/sample_060.jsonl
% Uni-OPD: /mnt/shared-storage-gpfs2/ai4energy/menglinjian/ckpts/opd-filter/
%      eval/Qwen3-4B-Base_from_GSPO-Qwen3-8B-Base_gbs960_
%      Uni-OPD_global_step_60/aime2024/sample_032.jsonl
% RA-OPD: eval/Qwen3-4B-Base_from_merged_hf_fF_fbtF_resp8192_k1_sp1_
%      dmk1T_dmodemask_global_step_100/aime2024/sample_056.jsonl
% ---------------------------------------------------------------------------
\subsection{The Case of Qwen3-4B-Base Student Model}
\label{app:case-qwen4b}

We first examine a representative mathematical reasoning case produced by the
Qwen3-4B-Base student model.

\noindent\textbf{Question.}
Let $\triangle ABC$ have circumcenter $O$ and incenter $I$ with
$\overline{IA}\perp\overline{OI}$, circumradius $13$, and inradius $6$.
Find $AB\cdot AC$.
\par\noindent\textbf{Reference answer:} $468$.

\begin{casepair}
\begin{rolloutbox}{OPD (Incorrect)}
\textbf{Model rollout (excerpt).}
``First, I need to understand what's being asked.''  The response recalls
Euler's identity and correctly computes
\[
OI^2=R^2-2Rr=13^2-2\cdot13\cdot6=13.
\]
It places $O$ at the origin and obtains from $IA\perp OI$
\[
(a-x)x-y^2=0,\qquad ax=x^2+y^2=13.
\]
It then introduces coordinates for $B,C$, calls the system
``complicated,'' and changes direction to area identities, deriving
\[
abc=4Rrs=312s.
\]
However, it never combines this with the perpendicularity.  It incorrectly
invokes a circle with diameter $OI$, then records
$AB=2R\sin C$ and $AC=2R\sin B$, hence
$AB\cdot AC=4R^2\sin B\sin C$, but cannot determine the remaining factor.

\reasoningomitted

After revisiting the same formulas without closing the system, it says,
``I'll make an educated guess. Suppose $AB\cdot AC=156$,'' without
verification.

\smallskip
\textbf{Model answer:} $156$.
\end{rolloutbox}
\begin{rolloutbox}{ExOPD (Incorrect)}
\textbf{Model rollout (excerpt).}
The response begins from the same valid calculation,
\[
OI^2=13^2-2\cdot13\cdot6=13.
\]
It expands the coordinate condition for $IA\perp OI$, calls it
``complicated,'' and switches to the sine rule:
\[
AB=2R\sin C,\qquad AC=2R\sin B,
\]
so the product is $4R^2\sin B\sin C$.  Unable to determine
$\sin B\sin C$, it abandons this route.

Next it recalls $IA=r/\sin(A/2)$ and the area identities
$\Delta=rs=abc/(4R)$, but never joins them.  After another detour it
silently replaces the correct product by
\[
AB\cdot AC=2R^2\sin A=338\sin A.
\]

\reasoningomitted

Its ``final attempt'' asserts without derivation that the perpendicularity
implies $A=90^\circ$.  Substitution into the already incorrect formula
gives $338$, with no geometric check.

\smallskip
\textbf{Model answer:} $338$.
\end{rolloutbox}
\end{casepair}

\begin{casepair}
\begin{rolloutbox}{Uni-OPD (Incorrect)}
\textbf{Model rollout (excerpt).}
The response first uses the area formulas
\[
\frac{abc}{4R}=rs
\quad\Longrightarrow\quad
abc=312s.
\]
It then places $O=(0,0)$ and $A=(13,0)$, lets $I=(x,y)$, and translates
the perpendicularity correctly:
\[
(13-x)x-y^2=0.
\]
Together with $x^2+y^2=OI^2=13$, this implies $x=1$, but the response does
not exploit it.  It introduces $B,C$ coordinates, finds the side-distance
conditions too involved, and switches to trigonometry.

It next confuses the angle at $O$ in triangle $OBC$ with vertex angle $A$.
It repeatedly returns to $abc=312s$ and moves among the law of cosines,
the inradius, and the vector condition without completing any route.

\reasoningomitted

It finally acknowledges that no direct relationship was found, but selects
$AB\cdot AC=169$.  This is merely $R^2$ and is unsupported.

\smallskip
\textbf{Model answer:} $169$.
\end{rolloutbox}
\begin{rolloutbox}[colback=green!3,colframe=green!45!black]{RA-OPD (Correct)}
\textbf{Model rollout (excerpt).}
The response computes $OI^2=13$ and $abc=312s$, then returns to the
geometric meaning of the perpendicularity: triangle $OIA$ is right at
$I$.  Hence
\[
AI^2=OA^2-OI^2=169-13=156.
\]
Writing $a=BC$, $b=CA$, $c=AB$, it recalls
\[
AI^2=\frac{bc(s-a)}{s},
\]
and combines it with
\[
abc=4R\Delta=4Rrs=312s.
\]

\reasoningomitted

For $x=bc=AB\cdot AC$, $ax=312s$ gives $a=312s/x$.  Substitution yields
\[
156=\frac{x(s-a)}{s}
   =x-\frac{xa}{s}
   =x-312.
\]
Thus $x=468$, consistently linked to both radii and the perpendicularity.

\smallskip
\textbf{Model answer:} $468$.
\end{rolloutbox}
\end{casepair}

\noindent\textbf{Comparison.}
All three baselines recover at most isolated ingredients and then introduce
an unsupported identity or guess.  RA-OPD connects the perpendicularity,
the vertex--incenter distance identity, and the area formula, which reduces
the requested product to a one-variable equation.

\clearpage

% ---------------------------------------------------------------------------
% Qwen3-8B-Base sources (all line 17 of the indicated JSONL):
% OPD: /mnt/shared-storage-gpfs2/ai4energy/menglinjian/ckpts/opd-filter/
%      eval/Qwen3-8B-Base_from_merged_hf_fF_fbtF_resp8192_k1_sp1_
%      dmk1F_dmodeza_global_step_100/aime2024/sample_028.jsonl
% ExOPD (legacy checkpoint directory named G-OPD): same root,
%      Qwen3-8B-Base_from_GSPO-Qwen3-8B-Base_gbs240_
%      resp8192_G-OPD_global_step_100/aime2024/sample_033.jsonl
% Uni-OPD: same root, Qwen3-8B-Base_from_GSPO-Qwen3-8B-Base_gbs960_
%      Uni-OPD_global_step_60/aime2024/sample_018.jsonl
% RA-OPD: same root, Qwen3-8B-Base_from_merged_hf_fF_fbtF_resp8192_k1_
%      sp1_dmk1T_dmodemask_global_step_100/aime2024/sample_006.jsonl
% ---------------------------------------------------------------------------
\subsection{The Case of Qwen3-8B-Base Student Model}
\label{app:case-qwen8b}

We next examine a representative mathematical reasoning case produced by the
Qwen3-8B-Base student model.

\noindent\textbf{Question.}
Let $p$ be the least prime for which $p^2\mid n^4+1$ for some positive
integer $n$.  Find the least positive integer $m$ such that
$p^2\mid m^4+1$.
\par\noindent\textbf{Reference answer:} $110$.

\begin{casepair}
\begin{rolloutbox}{OPD (Incorrect)}
\textbf{Model rollout (excerpt).}
The response first translates the task to
$n^4\equiv-1\pmod {p^2}$ and tests small primes.  For $p=2$ it correctly
observes that a fourth power is $0$ or $1$ modulo $4$.  It then lists
several residues modulo $9$ and $25$, finding no candidate for $p=3$ or
$p=5$.  At $p=7$ it stops the enumeration and says, ``perhaps I need a
better approach.''

The response remembers that $-1$ being a quadratic residue requires
$p\equiv1\pmod4$, tests a few values for $p=13$, and then jumps to:
``I recall that the smallest prime is $p=17$.''  No argument excluding
$p=13$ modulo $13^2$ or establishing the stronger fourth-power condition
is completed.

For the second part it recognizes that it must solve
$m^4\equiv-1\pmod {289}$ and calls such an $m$ a primitive eighth root.
Instead of lifting a residue, it says, ``I recall that for $p=17$, the
smallest $m$ is $130$.''  It then tries to verify the recalled value.  Its
own arithmetic gives
\[
130\equiv11\pmod {17},
\qquad 11^4\equiv4\pmod {17},
\]
which means $130^4+1\equiv5\pmod {17}$.

\reasoningomitted

Rather than rejecting the candidate after this contradiction, the response
says the modulo-$17$ computation ``doesn't directly help,'' calls the
modulo-$289$ calculation tedious, and concludes, ``Given the complexity,
I'll assume $m=130$ is the smallest.''  Thus the final answer conflicts
with the response's own check.

\smallskip
\textbf{Model answer:} $130$.
\end{rolloutbox}
\begin{rolloutbox}{ExOPD (Incorrect)}
\textbf{Model rollout (excerpt).}
The response also starts by testing primes directly.  It evaluates a few
small $n$ for $p=2$, $3$, and $5$, but does not enumerate complete residue
classes.  When the checks become lengthy, it says, ``This seems like a
problem involving modular arithmetic and number theory,'' followed by
``perhaps $p=17$ is the smallest prime where this holds.''  The assumption
is carried forward without proof.

For $p=17$, it correctly states the target
\[
m^4\equiv-1\pmod {289}.
\]
It starts a brute-force search with $m=1,2,3$ and immediately notes that
continuing this way ``will take a while.''  It mentions that a smarter
method is needed, but never develops Hensel lifting or a group-theoretic
construction.

\reasoningomitted

The response then abruptly proposes $m=44$: ``Given the complexity, I'll
assume that the smallest $m$ is $44$, based on known results or patterns.''
It neither evaluates $44^4+1$ modulo $289$ nor checks whether a smaller
solution exists.  In fact, $44\equiv10\pmod {17}$ and
$10^4\equiv4\pmod {17}$, so the candidate already fails modulo $17$.

\smallskip
\textbf{Model answer:} $44$.
\end{rolloutbox}
\end{casepair}

\begin{casepair}
\begin{rolloutbox}{Uni-OPD (Incorrect)}
\textbf{Model rollout (excerpt).}
The response explicitly plans to find the least prime first and the least
lifted integer second.  It correctly rules out $p=2$ and tabulates fourth
powers modulo $9$ for $p=3$.  It then turns to $p=5$ and says that testing
all residues modulo $25$ would be time-consuming.

At this point it makes the decisive arithmetic error:
\[
7^4=2401\equiv-1\pmod {25}.
\]
In fact, $2401=25\cdot96+1$, so the residue is $1$, not $24$.  Based on
this mistaken line, the response declares $p=5$ to be the least eligible
prime.

It next searches for the least $m$ modulo $25$.  Values $m=1$ through $6$
are checked one by one.  The response obtains residues $1$, $16$, $6$,
$6$, $0$, and $21$, none equal to $-1$, and therefore returns to the same
incorrectly evaluated candidate $m=7$.

\reasoningomitted

The final paragraph repeats that ``from our earlier testing, $m=7$
satisfies this'' and reports $7$.  The overall structure of the search is
reasonable, but the single false congruence invalidates both the chosen
prime and the chosen integer.

\smallskip
\textbf{Model answer:} $7$.
\end{rolloutbox}
\begin{rolloutbox}[colback=green!3,colframe=green!45!black]{RA-OPD (Correct)}
\textbf{Model rollout (excerpt).}
The response initially follows the same direct-testing route.  It rules out
$p=2$ and $p=3$, tries several residues modulo $25$, and briefly becomes
sidetracked by checking $2^4+1=17$, which is divisible by $17$ but not
$17^2$.  It then corrects course: ``Maybe I need a different approach.''

The response observes that a solution has order $8$ modulo $p$, so it
returns to $p=17$ and enumerates fourth powers modulo $17$.  The key
revision occurs at
\[
8^4=4096\equiv-1\pmod {17}.
\]
It now lifts this valid residue rather than guessing a value modulo
$17^2$.  Writing $m=8+17k$, the binomial expansion is truncated correctly
modulo $289$:
\[
(8+17k)^4\equiv50+136k\pmod {289}.
\]
The required congruence becomes
\begin{align*}
50+136k&\equiv-1\pmod {289},\\
8k&\equiv14\pmod {17}.
\end{align*}

\reasoningomitted

The response checks that $8^{-1}\equiv15\pmod {17}$ and obtains
$k\equiv14\cdot15\equiv6\pmod {17}$.  It therefore constructs
\[
m=8+17\cdot6=110.
\]
Unlike the baseline candidates, this number is produced by a valid
congruence lift; the response retains $110$ consistently in its concluding
line.

\smallskip
\textbf{Model answer:} $110$.
\end{rolloutbox}
\end{casepair}

\noindent\textbf{Comparison.}
OPD and ExOPD guess candidates that are not supported by the required
congruence, while Uni-OPD makes a direct modular-arithmetic error.  RA-OPD
performs the lift from a valid root modulo $17$ to a valid root modulo
$17^2$ and reaches the reference answer.

\clearpage

% ---------------------------------------------------------------------------
% DeepSeek-R1-Distill-Qwen-7B sources (all line 24):
% OPD: eval/DeepSeek-R1-Distill-Qwen-7B_from_Skywork-OR1-Math-7B_
%      fF_fbtF_resp16384_k1_sp1_dmk1F_dmodeza_global_step_100/
%      aime2024/sample_039.jsonl
% ExOPD (legacy checkpoint directory named G-OPD):
% /mnt/shared-storage-gpfs2/ai4energy/menglinjian/ckpts/opd-filter/
%      eval/DeepSeek-R1-Distill-Qwen-7B_from_Skywork-OR1-Math-7B_
%      gbs240_resp16384_G-OPD_global_step_100/aime2024/sample_002.jsonl
% Uni-OPD: same root, ..._gbs960_resp16384_Uni-OPD_global_step_100/
%      aime2024/sample_023.jsonl
% RA-OPD: eval/DeepSeek-R1-Distill-Qwen-7B_from_Skywork-OR1-Math-7B_
%      fF_fbtF_resp16384_k1_sp1_dmk1T_dmodemask_global_step_100/
%      aime2024/sample_039.jsonl
% ---------------------------------------------------------------------------
\subsection{The Case of DeepSeek-R1-Distill-Qwen-7B Student Model}
\label{app:case-deepseek7b}

We then examine a representative mathematical reasoning case produced by the
DeepSeek-R1-Distill-Qwen-7B student model.

\noindent\textbf{Question.}
Eight circles of radius $34$ are sequentially tangent, with the two end
circles tangent to $AB$ and $BC$ of triangle $ABC$, respectively.
The same arrangement can be formed with $2024$ circles of radius $1$.
If the inradius is $m/n$ in lowest terms, find $m+n$.
\par\noindent\textbf{Reference answer:} $197$.

\begin{casepair}
\begin{rolloutbox}{OPD (Incorrect)}
\textbf{Model rollout (excerpt).}
The response first debates whether every circle touches both sides or only
the endpoints do.  It imagines a geometric progression, then corrects
itself after noticing that all eight radii equal $34$: ``perhaps the
circles form a row from $AB$ to $BC$.''  It never turns this correction
into a consistent diagram.

The response nevertheless introduces a common ratio $k$ and assumes that
the inradius is the sum of a geometric sequence of radii:
\[
r=34(1+k+\cdots+k^7)
  =1+k+\cdots+k^{2023}.
\]
After canceling the geometric-series denominator it obtains
\[
34(k^8-1)=k^{2024}-1.
\]
It rejects $k=1$ because the arrangements use different radii, then tries
$k^8=34$, $k=17$, $\sqrt{17}$, and $2$, observing that the resulting
$k^{2024}$ is far too large.

\reasoningomitted

It then guesses from $34\cdot8$, $2024/8$, and $34\cdot3$.  ``Perhaps the
inradius is $102$'' has no construction behind it; it finally reports
$m+n=103$ and even calls $103$ the inradius.

\smallskip
\textbf{Model answer:} $103$.
\end{rolloutbox}
\begin{rolloutbox}{ExOPD (Incorrect)}
\textbf{Model rollout (excerpt).}
The response debates the picture, then assumes without basis that $ABC$ is
right.  Its proposed $a+b-c$ relation immediately gives incompatible
values:
\begin{align*}
a+b-c&=8(2\cdot34)=544,\\
a+b-c&=2024(2\cdot1)=4048.
\end{align*}
It correctly rejects that relationship.

It next rejects the false scaling $2024=8\cdot34$, then cycles through
radius sums, the ratio $2024/8=253$, candidates $8502$, $8503$, $287$,
and a harmonic mean, without connecting any to the triangle.

\reasoningomitted

Finally, the response proposes a new expression with no derivation from
tangency or center distances:
\[
r=34\frac{2024}{2024-8}
  =\frac{4301}{126}.
\]
It reports the coprime numerator and denominator's sum, but the formula
accounts for neither the $N-1$ center gaps nor the endpoint offsets.

\smallskip
\textbf{Model answer:} $4427$.
\end{rolloutbox}
\end{casepair}

\begin{casepair}
\begin{rolloutbox}{Uni-OPD (Incorrect)}
\textbf{Model rollout (excerpt).}
The response rejects a right-triangle assumption and naive radius scaling,
then makes a substantive correction: tangent centers are $2\rho$ apart.
It counts
\[
7\cdot68=476,
\]
and for $2024$ unit circles it counts
$2023\cdot2=4046$.

With $u=\sin(\theta/2)$, it uses endpoint offsets $\rho/u$, but models the
full span with only one inradius contribution, obtaining
\[
r=68+476u,\qquad r-2=4046u.
\]

\reasoningomitted

It first reports $388/5$, notices an inconsistency, and recomputes:
\[
3570u=66,\qquad u=\frac{11}{595},
\]
\[
r=68+476\left(\frac{11}{595}\right)=\frac{384}{5}.
\]
The arithmetic correction is valid, but the equations lost a factor of
two: the full span uses $2r/u$, not $r/u$.

\smallskip
\textbf{Model answer:} $389$.
\end{rolloutbox}
\begin{rolloutbox}[colback=green!3,colframe=green!45!black]{RA-OPD (Correct)}
\textbf{Model rollout (excerpt).}
The response initially tests an angle-bisector chain, a geometric sequence,
and sums of radii.  It rejects them after $8\cdot34$ and $2024\cdot1$
cannot describe one fixed triangle length.

\reasoningomitted

The decisive revision uses $(N-1)2\rho$ between the first and last centers
and retains offsets at \emph{both} ends.  The two chains therefore
contribute $2\cdot34/u+7\cdot68$ and $2/u+2023\cdot2$.

With $u=\sin(\theta/2)$ and inradius $r$, multiplying the two full-span
relations by $u$ gives
\[
68+476u=2r,\qquad 2+4046u=2r.
\]
The response subtracts the equations:
\[
u=\frac{66}{3570}=\frac{11}{595}.
\]
It then substitutes back and simplifies carefully:
\[
2r=68+476\left(\frac{11}{595}\right)
   =\frac{6528}{85},
\qquad r=\frac{192}{5}.
\]
It checks coprimality and reports $m+n=192+5=197$.

\smallskip
\textbf{Model answer:} $197$.
\end{rolloutbox}
\end{casepair}

\noindent\textbf{Comparison.}
The baselines either impose an unsupported scaling rule or lose one of the
two endpoint contributions.  RA-OPD uses the same tangent-center gap counts
as Uni-OPD but preserves the full span, eliminating the factor-of-two error.

\clearpage

% ---------------------------------------------------------------------------
% Code sources, title "closest-equal-element-queries", question_id 3750.
% All files are under /mnt/shared-storage-gpfs2/ai4energy/menglinjian/ckpts/
% opd_topk/code_eval/<run>/livecodebench_two_stage/lcb_outputs/huggingface/
% Scenario.codegeneration_16_0.6_eval_all.json.
% OPD run: Qwen3-4B-Base_from_huggingface_fF_fbtF_resp8192_k1_sp1_
%      dmk1F_dmodeza_global_step_200, rollout index 8.
% ExOPD run (legacy checkpoint directory named G-OPD):
%      Qwen3-4B-Base_from_GSPO-Qwen3-8B-Base-Eurus2-Code-25K_
%      gbs240_resp8192_G-OPD_global_step_200, rollout index 2.
% Uni-OPD run: Qwen3-4B-Base_from_GSPO-Qwen3-8B-Base-Eurus2-Code-25K-
%      step180_gbs960_Uni-OPD_global_step_60, rollout index 0.
% RA-OPD run: Qwen3-4B-Base_from_huggingface_fF_fbtF_resp8192_k1_sp1_
%      dmk1T_dmodemask_global_step_200, rollout index 8.
% ---------------------------------------------------------------------------
\subsection{The Case of Qwen3-4B-Base Code Student Model}
\label{app:case-qwen4b-code}

Finally, we examine a representative code generation case produced by the
Qwen3-4B-Base student model.

\noindent\textbf{Question.}
Given a circular array \texttt{nums} and query indices \texttt{queries},
return, for each queried index $q$, the minimum circular distance to a
different index $j$ with $\texttt{nums}[j]=\texttt{nums}[q]$; return $-1$
when no such $j$ exists.  For example,
\[
\texttt{nums}=[1,3,1,4,1,3,2],\quad
\texttt{queries}=[0,3,5]
\]
must produce $[2,-1,3]$.
\par\noindent\textbf{Reference criterion:} pass the LCB judge.

\begin{casepair}
\begin{rolloutbox}{OPD (Incorrect)}
\textbf{Model rollout (excerpt).}
The response first identifies the correct circular-distance formula:
\[
d(i,j)=\min\bigl(|i-j|,N-|i-j|\bigr).
\]
Its stated plan is to process each query, collect all positions containing
the queried value, compute this distance to each position, and return the
minimum.  It also says that $-1$ should be returned when no alternative
position exists.

In the implementation, however, the collection step includes the queried
index itself:
\[
\texttt{matching\_indices}
=\{i:\texttt{nums}[i]=\texttt{nums}[q]\}.
\]
The subsequent loop has no condition excluding $i=q$.  Therefore every
query immediately supplies the candidate $d(q,q)=0$, which is necessarily
the minimum.  The preceding empty-list test also cannot detect a unique
value, since the query position is always in the list.  Although the prose
repeatedly says ``any other index,'' that requirement is lost in the code.
On the public example, the implementation returns $[0,0,0]$ instead of
$[2,-1,3]$.

\smallskip
\textbf{Judge result:} wrong answer.
\end{rolloutbox}
\begin{rolloutbox}{ExOPD (Incorrect)}
\textbf{Model rollout (excerpt).}
The response follows a nearly correct four-step plan: identify the queried
value, collect its indices, compute
$\min(|q-j|,N-|q-j|)$ for every matching position, and return $-1$ if no
other position exists.  Unlike OPD, its loop explicitly contains
\[
\texttt{if j == query: continue}.
\]
Thus repeated values receive correct distances.

The singleton check is performed too early, however.  The code asks whether
the collected index list is empty \emph{before} removing $q$:
\[
\texttt{if not indices: return -1}.
\]
For a value that appears once, \texttt{indices} still contains $q$, so the
condition is false.  The loop then skips its only member, leaving
\texttt{min\_distance} equal to infinity, which is appended directly.
The prose claims the code returns $-1$ in this case, but there is no
post-loop infinity check.  Consequently the public query at index $3$,
whose value $4$ is unique, produces \texttt{inf}; the complete public
output is $[2,\texttt{inf},3]$.

\smallskip
\textbf{Judge result:} wrong answer.
\end{rolloutbox}
\end{casepair}

\begin{casepair}
\begin{rolloutbox}{Uni-OPD (Incorrect)}
\textbf{Model rollout (excerpt).}
The response correctly excludes the queried index and maintains a
\texttt{found} flag so that singleton values produce $-1$.  Its high-level
description also recognizes that the array is circular.  The error lies in
how it converts index order into distance.  For each matching $j$, it uses
\[
d(q,j)=
\begin{cases}
|q-j|, & q<j,\\
N-|q-j|, & q\ge j.
\end{cases}
\]
This selects the ordinary forward gap when the match is to the right and
the wraparound gap when it is to the left.  Circular distance is not
determined by that ordering: both directions must be compared for every
pair.

For example, if $N=18$, $q=13$, and a matching position is $j=5$, the
response uses $18-|13-5|=10$ and never considers the shorter ordinary gap
$8$.  The same omission causes several hidden-test outputs to be too large.
Thus the surrounding control flow is more careful than the other
baselines, but the core distance expression still implements only one of
the two circular directions.

\smallskip
\textbf{Judge result:} wrong answer.
\end{rolloutbox}
\begin{rolloutbox}[colback=green!3,colframe=green!45!black]{RA-OPD (Correct)}
\textbf{Model rollout (excerpt).}
The response separates preprocessing from query handling.  It first maps
each value to all of its positions:
\[
\texttt{value\_to\_indices[value].append(index)}.
\]
For a query $q$, it retrieves only the positions sharing
\texttt{nums[q]}, initializes the running minimum to infinity, and
explicitly skips $j=q$.  For every remaining candidate it computes both
directions:
\[
\delta=|j-q|,
\qquad d(q,j)=\min(\delta,N-\delta),
\]
then takes the minimum over all matching positions.

The response also handles the edge case that the baselines miss.  After the
loop, infinity means that no distinct matching position remained, so it
appends $-1$; otherwise it appends the finite minimum.  On the public
example this gives $2$ for the nearest repeated $1$, $-1$ for the unique
$4$, and $3$ for the repeated $3$ through the wraparound path.  The judged
implementation therefore agrees with both the public behavior and the
hidden tests used for this rollout.

\smallskip
\textbf{Judge result:} passed.
\end{rolloutbox}
\end{casepair}

\noindent\textbf{Comparison.}
The three baseline implementations respectively include the query itself,
mishandle singleton values, or consider only one circular direction.
RA-OPD satisfies all three requirements.  On this prompt, none of the
$16$ judged responses from OPD, ExOPD, or Uni-OPD passes, while $2$ of the
$16$ RA-OPD responses pass.

\end{document}